\pdfoutput=1

\documentclass[11pt]{article}
\usepackage[final]{acl}

\usepackage{times}
\usepackage{latexsym}

\usepackage[T1]{fontenc}

\usepackage[utf8]{inputenc}

\usepackage{microtype}

\usepackage{inconsolata}

\usepackage{graphicx}

\usepackage{colortbl}
\usepackage{hyperref}
\usepackage{url}
\usepackage{kotex}
\usepackage{amsmath}
\usepackage{amssymb}
\usepackage{multirow}
\usepackage{adjustbox}
\usepackage{makecell}
\usepackage{booktabs}  

\title{Making Sense of Korean Sentences:\\A Comprehensive Evaluation of LLMs through KoSEnd Dataset}

\author{Seunguk Yu, Kyeonghyun Kim, Jungmin Yun \and Youngbin Kim \\
  Chung-Ang University, Seoul, Republic of Korea \\
  \texttt{seungukyu@gmail.com, \{khyun8072, cocoro357, ybkim85\}@cau.ac.kr} \\
}

\begin{document}
\maketitle
\begin{abstract}
Although LLMs have made significant progress in various languages, there are still concerns about their effectiveness with low-resource agglutinative languages compared to languages such as English. In this study, we focused on Korean, a language known for its complex sentence endings, and evaluated LLMs on this challenging aspect. We introduce the Korean Sentence Endings (\textbf{KoSEnd}) dataset, which includes 3,000 sentences, each annotated for the naturalness of 15 sentence ending forms. These were collected from diverse sources to cover a range of contexts. We evaluated 11 LLMs to assess their understanding of Korean sentence endings, analyzing them based on parameter count and prediction consistency. Notably, we found that informing models about the possibility of missing sentence endings improved performance, highlighting the impact of explicitly considering certain linguistic features.
\end{abstract}

\section{Introduction}

With the continuous advancement of large language models (LLMs), they have become capable of understanding multiple languages, irrespective of the input language~\cite{zhang2023m3exam, huang-etal-2023-languages}. However, the data used to train these models are heavily skewed toward English, rather than being evenly distributed across various languages~\cite{liu2024translation, li2024quantifying}. Consequently, LLMs may exhibit varying levels of comprehension depending on the language used, raising concerns regarding their effectiveness in understanding relatively low-resource languages~\cite{cahyawijaya-etal-2024-llms, asai-etal-2024-buffet, cahyawijaya-etal-2023-nusacrowd}.

Moreover, languages with alphabetic scripts often have advantages in tokenization since they can share some of the model's limited token capacity~\cite{petrov2024language, limisiewicz-etal-2023-tokenization}, while non-alphabetic script languages often face challenges due to smaller training datasets. Additionally, agglutinative languages like Korean have complex morphological structures, which further complicate tokenization and related processes~\cite{song2024multilingual, kaya2024effect}. Consequently, LLMs tend to be disproportionately advantaged in alphabetic languages compared to relatively low-resource agglutinative languages.

\begin{figure}[t!]
    \centerline{\includegraphics[width=\columnwidth]{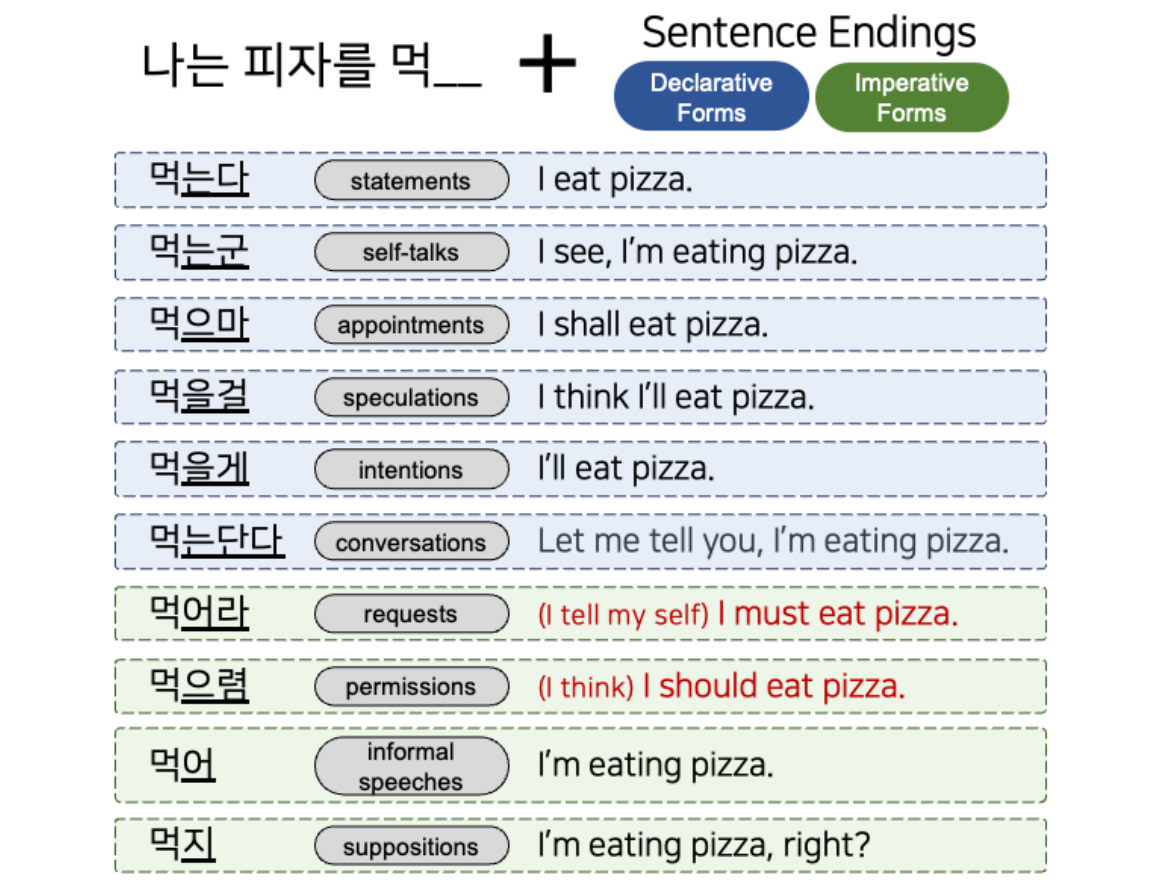}}
    \caption{Impact of the Korean sentence endings on the meaning of sentences. The translated texts showed that even small differences in sentence endings can lead to significant changes in meaning.}
    \label{figure_sentence_endings} 
\end{figure}

\begin{figure*}[t!]
    \centerline{\includegraphics[width=\textwidth]{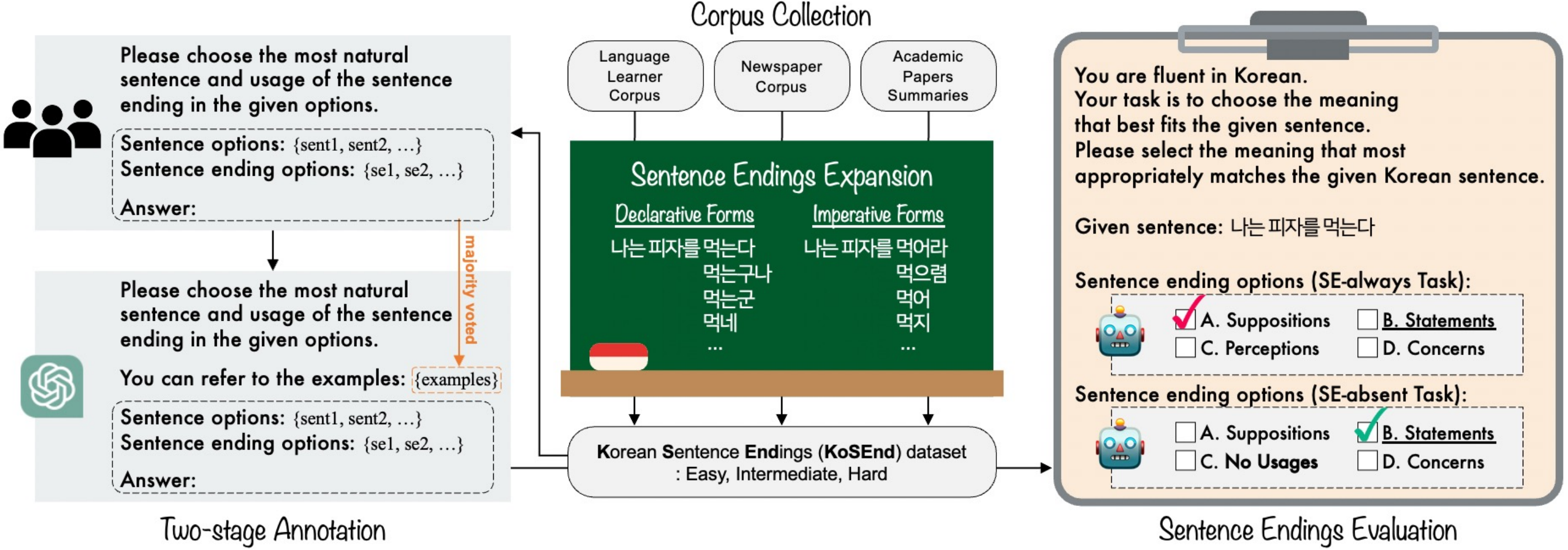}}
    \caption{Process of constructing the Korean Sentence Endings (\textbf{KoSEnd}) dataset and evaluating LLMs' understanding of Korean sentence endings. Sections \textsection\ref{section_3.1} and \textsection\ref{section_3.2} cover the \textit{Corpus Collection} and \textit{Sentence Ending Expansion}, respectively. Section \textsection\ref{section_3.3} describes the \textit{Two-stage Annotation}, and these three sections constitute the process of constructing the dataset. Section \textsection\ref{section_4} presents the \textit{Sentence Ending Tasks}, where we evaluated the LLMs understanding in Korean sentence endings through the designed tasks.}
    \label{figure_pipeline} 
\end{figure*}

In this case, we focus on the Korean language with agglutinative characteristics~\cite{sohn2001korean}. In Korean, a single verb stem can be combined with various sentence endings to express different meanings such as \textit{statements}, \textit{perceptions}, and \textit{exclamations}~\cite{lee2005korean}. As illustrated in Figure~\ref{figure_sentence_endings}, minor changes in sentence endings can significantly affect a sentence's meaning and interpretation\footnote{\label{note1}When using translation tools such as Google Translate or DeepL, we found that they fail to capture the nuances of Korean sentence endings accurately. To address this, we instructed the latest \texttt{gpt-4o} model to perform zero-shot translation with careful attention to the use of sentence endings.}. For example, while the blue expressions with \texttt{Declarative} endings generally convey the intended meanings, the green expressions with \texttt{Imperative} endings can feel awkward in certain contexts\footnote{In Figure~\ref{figure_sentence_endings}, some sentences may sound awkward as certain \texttt{Imperative} endings were used with the subject `\textit{I}.' These sentences are highlighted in red within the figure.}. This shows that sentence endings significantly impact the meaning and interpretation of a sentence, depending on the context.

Considering these perspectives, we examine the diverse usages of sentence endings and evaluate LLMs in this area. The construction of the proposed dataset and evaluation process are illustrated in Figure~\ref{figure_pipeline}. We propose the Korean Sentence Endings (\textbf{KoSEnd}) dataset, which explores the use of sentence endings in various contexts. Each sentence was expanded to include all theoretically possible sentence endings applicable to both \texttt{Declarative} and \texttt{Imperative} forms~\cite{lee2005korean}, ensuring that the dataset captures a wide range of contextual variations. Subsequently, we conducted a two-stage annotation process to reflect the natural usage of these endings based on context.

Using the proposed dataset, we evaluate how well various LLMs understand Korean sentence endings. We then analyze the results, taking into account factors such as model parameters and the consistency of their predictions. We found that each model had a different level of understanding of Korean sentence endings, with performance improving notably when we introduced the possibility that sentence endings \textit{might be absent}. Based on these results, and the observation that learning linguistic knowledge together contributed to improved performance on downstream tasks~\cite{xiang-etal-2022-visualizing, ke-etal-2020-sentilare, miaschi-etal-2020-linguistic}, we expect models with a deeper understanding of Korean sentence endings to also perform better on general tasks\footnote{We will publicly release the proposed dataset to encourage further research. \url{https://github.com/seungukyu/KoSEnd}}.

The contributions of our study are as follows:

\begin{itemize}
\item We propose the Korean Sentence Endings (\textbf{KoSEnd}) dataset, a collection of corpora categorized by the contextual difficulty. It includes sentence ending expansion and two-stage annotation process that capture the natural usages of Korean sentence endings.

\item We evaluate 11 LLMs to assess their understanding of Korean sentence endings. We compared performance by parameter count and analyzed prediction consistency across option orders, identifying models with robust comprehension of Korean sentence endings.

\item We further explore how informing models about the potential absence of sentence endings affected their performance. Across all models, performance improved with this consideration, suggesting that LLMs better grasp Korean sentence endings when considering this linguistic feature.
\end{itemize}

\section{Related Work}

\begin{table*}[t!]
\centering
\small
\begin{adjustbox}{max width=0.95\textwidth}
\begin{tabular}{l|l|l}
\hline
\rowcolor{gray!10}
\begin{tabular}[c]{@{}l@{}}Sentence Endings\\ in \texttt{Declarative} Forms\end{tabular} & Usages                                                                                                                                                                                         & Sentence Examples                                                                                                                                                   \\ \hline
\underline{\smash{(1)}} \{다, 는다, ㄴ다\}                                                                                & \textit{statements}, \textit{exclamations}, \textit{questions}                                                                                                                                 & \begin{tabular}[c]{@{}l@{}}보통 마음대로 좋은 선물을 가지고 간\underline{\smash{다}}\\ (They usually bring a good gift as they please.)\end{tabular}                                                    \\ \hline
\underline{\smash{(2)}} \{구나, 는구나\}                                                                                  & \textit{perceptions}, \textit{suppositions}                                                                                                                                                    & \begin{tabular}[c]{@{}l@{}}결말에 주인공이 국가를 위해 목숨을 바치는\underline{\smash{구나}}\\ (Ah, in the end, the main character sacrifices their life for the country.)\end{tabular}                     \\ \hline
\underline{\smash{(3)}} \{군, 는군\}                                                                                    & \textit{self-talks}, \textit{perceptions}                                                                                                                                                      & \begin{tabular}[c]{@{}l@{}}얘기를 많이 하니까 시간이 빨리 가\underline{\smash{는군}}\\ (Time sure flies when you talk a lot.)\end{tabular}                                                              \\ \hline
\underline{\smash{(4)}} \{네\}                                                                                        & \begin{tabular}[c]{@{}l@{}}\textit{perceptions}, \textit{exclamations},\\ \textit{self-talks}, \textit{questions}\end{tabular}                                                                 & \begin{tabular}[c]{@{}l@{}}그래서 우리는 학교 근처 편의점에 가\underline{\smash{네}}\\ (So, we ended up going to the convenience store near the school.)\end{tabular}                                   \\ \hline
\underline{\smash{(5)}} \{으마, 마\}                                                                                    & \textit{appointments}, \textit{intentions}                                                                                                                                                     & \begin{tabular}[c]{@{}l@{}}학생들이 잘 공부하도록 언제나 최선을 다하\underline{\smash{마}}\\ (I will always do my best so that the students can study well.)\end{tabular} \\ \hline
\underline{\smash{(6)}} \{을걸, 걸\}                                                                                    & \textit{speculations}                                                                                                                                                                          & \begin{tabular}[c]{@{}l@{}}벌써 1년이나 지났는데 지금 그날을 생각하면 아직도 행복한 느낌이 들\underline{\smash{걸}}\\ (It's already been a year, but when I think about that day, I still feel happy.)\end{tabular}  \\ \hline
\underline{\smash{(7)}} \{을게, ㄹ게, 을래, 래\}                                                                            & \begin{tabular}[c]{@{}l@{}}\textit{(expressions of) intentions},\\ \textit{questions}\end{tabular}                                                                                             & \begin{tabular}[c]{@{}l@{}}한국 문화에 관심이 있\underline{\smash{을래}}\\ (I think I might be interested in Korean culture.)\\ (Would you be interested in Korean culture?)\end{tabular}          \\ \hline
\underline{\smash{(8)}} \{을라, ㄹ라\}                                                                                   & \textit{concerns}                                                                                                                                                                              & \begin{tabular}[c]{@{}l@{}}많은 사람들이 물가가 너무 올라가서 걱정을 \underline{\smash{할라}}\\ (Many people are worried because the cost of living has gone up too much.)\end{tabular}                     \\ \hline
\underline{\smash{(9)}} \{는단다, ㄴ단다, 단다, 란다\}                                                                         & \textit{conversations}                                                                                                                                                                         & \begin{tabular}[c]{@{}l@{}}아주 힘들었지만 예쁜 경치를 봐서 기분이 좋\underline{\smash{단다}}\\ (It was really tough, but I feel good because I got to see the beautiful scenery.)\end{tabular}             \\ \hline
\rowcolor{gray!10}
\begin{tabular}[c]{@{}l@{}}Sentence Endings\\ in \texttt{Imperative} Forms\end{tabular}  & Usages                                                                                                                                                                                         & Sentence Examples                                                                                                                                                   \\ \hline
\underline{\smash{(10)}} \{아라, 어라, 여라\}                                                                               & \begin{tabular}[c]{@{}l@{}}\textit{commands}, \textit{requests},\\ \textit{permissions}, \textit{exclamations}\end{tabular}                                                                    & \begin{tabular}[c]{@{}l@{}}한국에서 간 장소에서 홍대를 소개하\underline{\smash{여라}}\\ (Introduce \textit{Hongdae} among the places you visited in Korea.)\end{tabular}                                          \\ \hline
\underline{\smash{(11)}} \{으려무나, 려무나, 으렴, 렴\}                                                                         & \textit{permissions}, \textit{commands}                                                                                                                                                        & \begin{tabular}[c]{@{}l@{}}돈을 벌고 나서 같이 여행하\underline{\smash{렴}}\\ (After you earn some money, let's go on a trip together.)\end{tabular}                                                \\ \hline
\underline{\smash{(12)}} \{소서\}                                                                                       & \textit{hopes}                                                                                                                                                                                 & \begin{tabular}[c]{@{}l@{}}장애인에게 많은 관심을 가지\underline{\smash{소서}}\\ (Please show a lot of interest in people with disabilities.)\end{tabular}                                            \\ \hline
\underline{\smash{(13)}} \{어\}                                                                                        & \textit{informal speeches}                                                                                                                                                                     & \begin{tabular}[c]{@{}l@{}}게다가 이 일을 하면 스트레스가 많\underline{\smash{어}}\\ (Besides, doing this job causes a lot of stress.)\end{tabular}                                                    \\ \hline
\underline{\smash{(14)}} \{아\}                                                                                        & \textit{informal speeches}, \textit{surprises}                                                                                                                                                 & \begin{tabular}[c]{@{}l@{}}명동은 사람이 많\underline{\smash{아}}\\ (\textit{Myeongdong} is crowded with people.)\end{tabular}                                                                           \\ \hline
\underline{\smash{(15)}} \{지\}                                                                                        & \begin{tabular}[c]{@{}l@{}}\textit{questions of confirmation},\\ \textit{obvious statements}, \textit{suppositions},\\ \textit{gentleness}, \textit{intentions}, \textit{regrets}\end{tabular} & \begin{tabular}[c]{@{}l@{}}나는 인생에 대한 새로운 생각이 생기\underline{\smash{지}}\\ (I've come to have new thoughts about life.)\end{tabular}                                                        \\ \hline
\end{tabular}
\end{adjustbox}
\caption{All forms of sentence endings used in this study, along with their usages and examples\footref{note1}~\cite{lee2005korean}. The top nine sentence ending forms are categorized as \texttt{Declarative}, while the bottom six are \texttt{Imperative}. Each ending is further grouped by usage, with the underlined Korean expressions in the `Sentence Examples' highlighting the specific endings used in each example.}
\label{table_sentence_endings}
\end{table*}

\subsection{NLP Benchmarks}

Numerous benchmarks have been developed to evaluate the reasoning abilities of language models. A notable research is SQuAD, which involves collecting question pairs for reading comprehension, along with its adaptations~\cite{rajpurkar-etal-2018-know, rajpurkar-etal-2016-squad}. Afterward, GLUE emerged with a broad set of language understanding tasks such as QA and natural language inference~\cite{wang-etal-2018-glue}. Subsequently, a method for evaluating the multitask performance of language models has been introduced, reflecting the ongoing research aimed at assessing model performance from multiple perspectives~\cite{bai-etal-2024-longbench, hendryckstest2021}.

Recently, several Korean natural language inference datasets have been developed using sources such as Wikipedia and news articles~\cite{park-etal-2021-klue, ham-etal-2020-kornli}. Research has progressed in utilizing linguistic features to understand sentence relationships~\cite{jang-etal-2022-kobest, lim2019korquad1} and measuring national alignment, particularly with the advanced LLMs~\cite{lee-etal-2024-kornat}. In this study, we construct an evaluation dataset grounded in the linguistic characteristics of the Korean language and conduct a comparative assessment of various LLMs.

\subsection{Commonsense Knowledge Evaluation}

Research on analytic languages, such as English, often struggles when applied to agglutinative languages with complex word formation. Recent studies reveal that LLMs face these challenges, highlighting the need for models that effectively address linguistic diversity~\cite{maxutov2024llms, weissweiler-etal-2023-counting}. In response, benchmarks have been introduced for natural language understanding tasks in agglutinative languages, including Japanese, Indonesian, and Kazakh~\cite{kurihara2022jglue, wilie-etal-2020-indonlu}.

Specifically, several datasets have been designed to evaluate the bias and dialogue comprehension of LLMs to assess their ability to understand nuanced semantic information in Korean~\cite{jang-etal-2024-kodialogbench, jin2024kobbq}. Nevertheless, performance comparisons from cultural and regional sources have noticed that LLMs encounter challenges in commonsense reasoning within a Korean-specific context~\cite{son2024kmmlu, son-etal-2024-hae, kim-etal-2024-click}.

\subsection{Linguistic Knowledge Evaluation}

Recent works have evaluated LLMs handling of morphological complexities and structural challenges in low-resource and agglutinative languages~\cite{nasution2024chatgpt, leong2023bhasa}. In Korean, studies have specifically examined the linguistic knowledge, including their understanding of grammatical structures and language proficiency~\cite{seo2024kocommongen}. For instance, studies analyzing linguistic factors, such as case markers and pragmatic competence, offer deeper insights into LLM performance in Korean~\cite{hwang2024kosmic, kim2024does, park2024pragmatic}. 

\section{KoSEnd: Dataset Construction}

\subsection{Corpus Collection}
\label{section_3.1}

Recognizing that Korean sentence endings can vary depending on the context, we collected three corpora, each categorized by the difficulty level: \texttt{Easy} from the language learner corpus, \texttt{Intermediate} from the newspaper corpus, and \texttt{Hard} from the academic papers summaries. The details regarding each corpus are provided in Appendix~\ref{appendix_a1}.

\subsection{Sentence Ending Expansion}
\label{section_3.2}

We expanded the original sentences from the corpora with diverse sentence endings. We focused on the \texttt{Declarative} and \texttt{Imperative} forms, which were categorized into nine and six types, as shown in Table~\ref{table_sentence_endings}. In Korean, sentence endings can be categorized into \texttt{Declarative}, \texttt{Interrogative}, and \texttt{Imperative} forms~\cite{lee2005korean}. For the \texttt{Interrogative} form, the presence of a question mark makes the use of specific endings straightforward. Therefore, we only focused on the endings used in \texttt{Declarative} and \texttt{Imperative} forms, which are more distinct and challenging.

The choice of appropriate sentence ending can be subjective, varying among readers based on their interpretation of context and communicative intent\footnote{Examples of unnatural sentence ending usage are provided in Appendix~\ref{appendix_a2}, depending on the context.}. Therefore, we conducted an annotation process to ensure the natural usages of sentence endings after expanding all sentences using a total of fifteen different sentence endings for \texttt{Declarative} and \texttt{Imperative} forms. The explanations of some examples in Table~\ref{table_sentence_endings} are explained in Appendix~\ref{appendix_a2}.

\subsection{Two-stage Annotation}
\label{section_3.3}

To establish standards for determining the natural use of sentence endings, we conducted a two-stage annotation process after expanding all the sentences. We began by performing human annotation on a subset of 20 sentences, covering 300 sentence ending instances from each difficulty level of the corpus. We found that even annotations from native Korean speakers can be inconsistent, as shown in Table~\ref{table_human_agreement_scores}. Given this situation, manually annotating the remaining sentences per difficulty level would be highly inefficient\footnote{It will require a total of 980$\times$15$\times$3$=$44,100 sentence ending cases for each, in terms of both time and cost.}. Therefore, for the cases not human-annotated, we utilized an LLM-based annotation~\cite{he-etal-2024-annollm, ding-etal-2023-gpt}.

\begin{table}[t!]
\begin{adjustbox}{max width=\columnwidth}
\centering
\small
\begin{tabular}{l|cc|cc}
\hline
\multirow{2}{*}{Difficulty} & \multicolumn{2}{c|}{\texttt{Declarative}} & \multicolumn{2}{c}{\texttt{Imperative}} \\ \cline{2-5} 
                            & Sentences           & Usages              & Sentences          & Usages             \\ \hline
\texttt{Easy}                 & 0.748               & \textbf{0.634}      & 0.733              & \textbf{0.644}     \\ \hline
\texttt{Intermediate}         & \textbf{0.755}      & 0.453               & \textbf{0.857}     & 0.544              \\ \hline
\texttt{Hard}                 & 0.556               & 0.300               & 0.594              & 0.417              \\ \hline
\end{tabular}
\end{adjustbox}
\caption{Krippendorff's $\alpha$~\cite{hayes2007answering} based on the human annotation results for each difficulty level. We found that easier levels resulted in higher scores and greater consistency among annotators, while scores decreased as difficulty increased, indicating more variation in the annotations.}
\label{table_human_agreement_scores}
\end{table}

\begin{table}[t!]
\begin{adjustbox}{max width=\columnwidth}
\centering
\small
\begin{tabular}{l|cc|cc}
\hline
\multirow{2}{*}{Difficulty}    & \multicolumn{2}{c|}{\texttt{Declarative}} & \multicolumn{2}{c}{\texttt{Imperative}} \\ \cline{2-5} 
                               & Sentences           & Usages              & Sentences          & Usages             \\ \hline
\texttt{Easy}                    & 53.69               & 64.62               & 54.99              & 54.99              \\ \hline
\texttt{Easy} (w/o \textit{None})         & 79.51               & 97.81               & 79.99              & 72.21              \\ \hline
\texttt{Intermediate}            & 77.58               & 91.10               & 50.55              & 53.60              \\ \hline
\texttt{Intermediate} (w/o \textit{None}) & 81.41               & 95.94               & 72.77              & 72.91              \\ \hline
\texttt{Hard}                    & 74.44               & 82.77               & 48.88              & 47.49              \\ \hline
\texttt{Hard} (w/o \textit{None})         & \textbf{87.58}      & \textbf{96.06}      & \textbf{80.41}     & \textbf{74.44}     \\ \hline
\end{tabular}
\end{adjustbox}
\caption{Accuracy on the model's classification with samples used for annotation. The gold labels were majority voted by the results among the annotators. The difficulty with (w/o \textit{None}) excludes samples where the gold label was labeled as \textit{None}.}
\label{table_model_annotation_acc}
\end{table}

\begin{table*}[t!]
\begin{adjustbox}{max width=\textwidth}
\centering
\begin{tabular}{l|ccc|c|c|cc|cc|c|c}
\hline
\multirow{2}{*}{Sentence Endings}                                                                     & \texttt{Llama3.1} & \texttt{Llama3} & \texttt{Llama3-ko}        & \texttt{KULLM3} & \texttt{EXAONE3} & \multicolumn{2}{c|}{\texttt{Qwen2}} & \multicolumn{2}{c|}{\texttt{Gemma2}} & \texttt{Openchat} & \texttt{Synatra} \\ \cline{2-12} 
                                                                                  & \multicolumn{3}{c|}{8B}                                         & 10.7B           & 7.8B             & 1.5B             & 7B               & 2B                          & 9B     & 8B                   & 7B                     \\ \hline
\multirow{3}{*}{\begin{tabular}[c]{@{}l@{}}\texttt{Declarative}\\ Forms\end{tabular}} & 13.06             & 15.09           & 17.33                     & 14.98           & 15.41            & 13.83            & 13.23            & 16.33                       & 14.44  & 13.49                & 16.64                  \\
                                                                                      & 13.47             & 17.23           & 20.14                     & 17.07           & 14.40            & 15.14            & 13.54            & 16.85                       & 13.83  & 14.18                & 16.84                  \\
                                                                                      & 12.33             & 15.77           & 18.31                     & 16.82           & 13.85            & 14.25            & 12.54            & 15.78                       & 13.05  & 13.35                & 15.46                  \\ \hline
Average                                                                               & 12.95             & 16.03           & \textbf{18.59}            & 16.29           & 14.55            & 14.40            & 13.10            & \underline{\smash{16.32}}   & 13.77  & 13.67                & 16.31                  \\ \hline
\multirow{3}{*}{\begin{tabular}[c]{@{}l@{}}\texttt{Imperative}\\ Forms\end{tabular}}  & 8.71              & 10.32           & 10.67                     & 10.28           & 9.49             & 10.47            & 8.79             & 9.68                        & 9.66   & 9.31                 & 10.97                  \\
                                                                                      & 8.67              & 12.40           & 12.26                     & 11.75           & 9.91             & 11.23            & 10.23            & 9.92                        & 10.66  & 10.65                & 12.16                  \\
                                                                                      & 8.43              & 11.02           & 11.33                     & 11.40           & 10.81            & 10.97            & 10.53            & 10.78                       & 11.35  & 10.39                & 11.70                  \\ \hline
Average                                                                               & 8.60              & 11.24           & \underline{\smash{11.42}} & 11.14           & 10.07            & 10.89            & 9.85             & 10.12                       & 10.55  & 10.11                & \textbf{11.61}         \\ \hline
\end{tabular}
\end{adjustbox}
\caption{Accuracy of understanding Korean sentence endings across LLMs \underline{\smash{for the SE-\textit{always} task}}. We determined each model's final accuracy using cyclic permutation, following the approach used in previous work~\cite{kim-etal-2024-click}. For both \texttt{Declarative} and \texttt{Imperative} forms, the three reported values from the top represent results for \texttt{Easy}, \texttt{Intermediate}, and \texttt{Hard}, respectively. The model with the highest average score across all models is highlighted in bold, whereas the second-best model is underlined.}
\label{table_SE_always}
\end{table*}

To evaluate whether the selected model efficiently understands Korean sentence endings, we provided it with the samples used for human annotation\footnote{In this case, we instructed the latest \texttt{gpt-4-turbo} model to perform zero-shot classification with careful attention to the use of sentence endings.}. We then compared the model's predictions to the majority voted human annotations and the accuracy results are provided in Table~\ref{table_model_annotation_acc}. The model achieved high accuracy in nearly all cases, generally aligning with the human annotation results.

Although the model demonstrated reliable performance, reaching a certain level of accuracy, we remained cautious about the potential for misclassifying sentence endings when annotating the remaining sentences. To address this, we employed following two strategies to enhance the model's ability to predict the usage of sentence endings accurately. First, we employed few-shot learning~\cite{brown2020language} by selecting a random sample of sentences and their sentence endings from human-annotated results that matched the usage patterns to predict. Second, we employed cyclic permutation~\cite{izacard2023atlas} when presenting options in the prompts to ensure unbiased model predictions independent of the order of the options, allowing it to focus on consistent patterns across different arrangements. The full annotation example and prompt configurations are provided in Appendix~\ref{appendix_a3}. Finally, we constructed a dataset that includes 1,000 sentences for each difficulty level with 15 different sentence endings applied to each sentence.

\section{Sentence Ending Tasks}

\label{section_4}

We defined specific tasks to evaluate LLMs' understanding of sentence endings by selecting the most contextually natural option from the provided choices for each sentence ending. As mentioned earlier, the appropriate usage of sentence endings depends on the context, and their natural application may be absent in some cases.

In this scenario, we evaluated model performance in two cases: one where a natural ending is always expected (SE-\textit{always}) and one where it may sometimes be absent (SE-\textit{absent})\footnote{In the following discussion of experimental results, we referred to the tasks as either SE-\textit{always} or SE-\textit{absent}, depending on which task was applied to evaluate the models.}. In the SE-\textit{always} task, we excluded samples labeled \textit{no usages} for each sentence ending and only included samples with labeled usages. In contrast, the SE-\textit{absent} task allowed \textit{no usages} as an option among the choices. This setup enabled us to compare model performance while considering the possibility of a missing natural sentence ending. For both tasks, we provided the model with four usage options for each sentence in a multiple-choice format. The details including the rules for presenting options are provided in Appendix~\ref{appendix_b1}.

\begin{figure}[t!]
    \centerline{\includegraphics[width=\columnwidth]{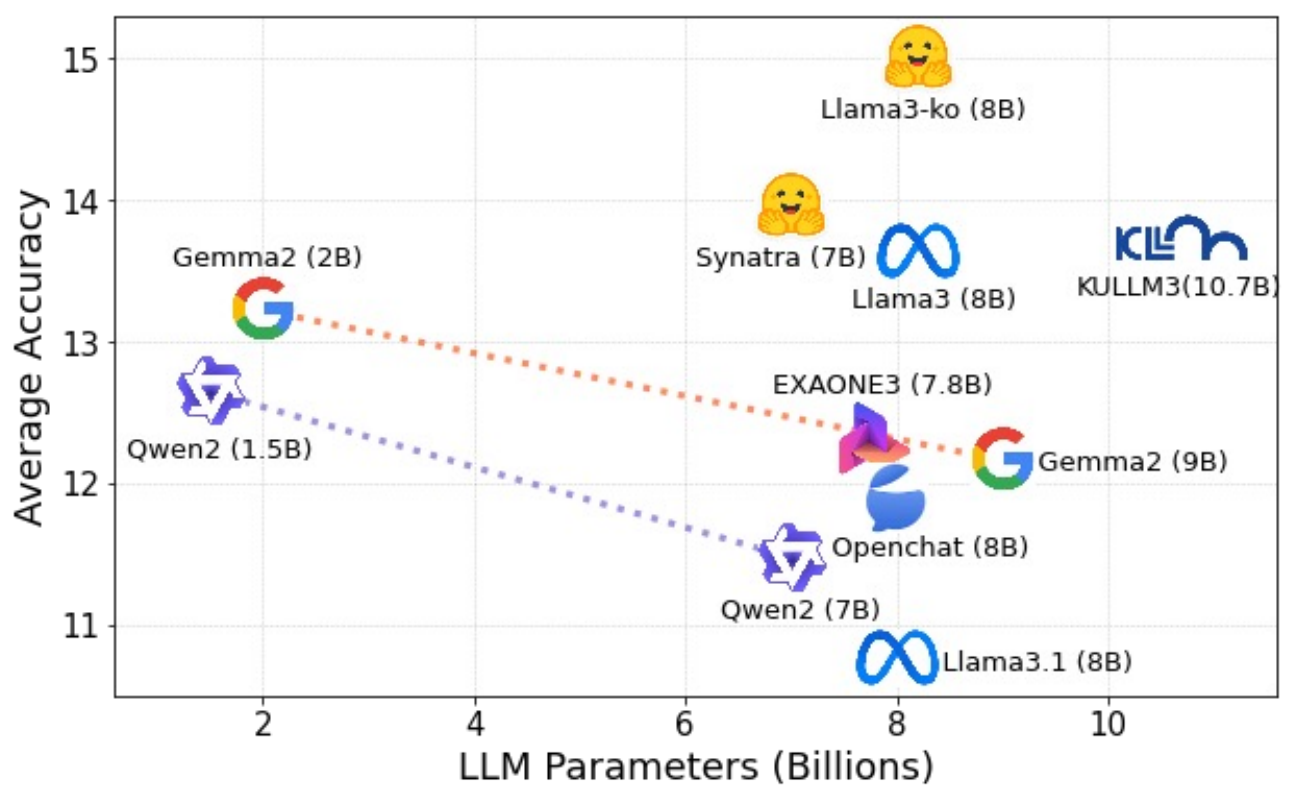}}
    \caption{Comparison across LLMs based on parameter count, with scores averaged over all six difficulty levels for both \texttt{Declarative} and \texttt{Imperative} forms.}
    \label{figure_compare_llms} 
\end{figure}

\begin{figure*}[t!]
    \centerline{\includegraphics[width=\textwidth]{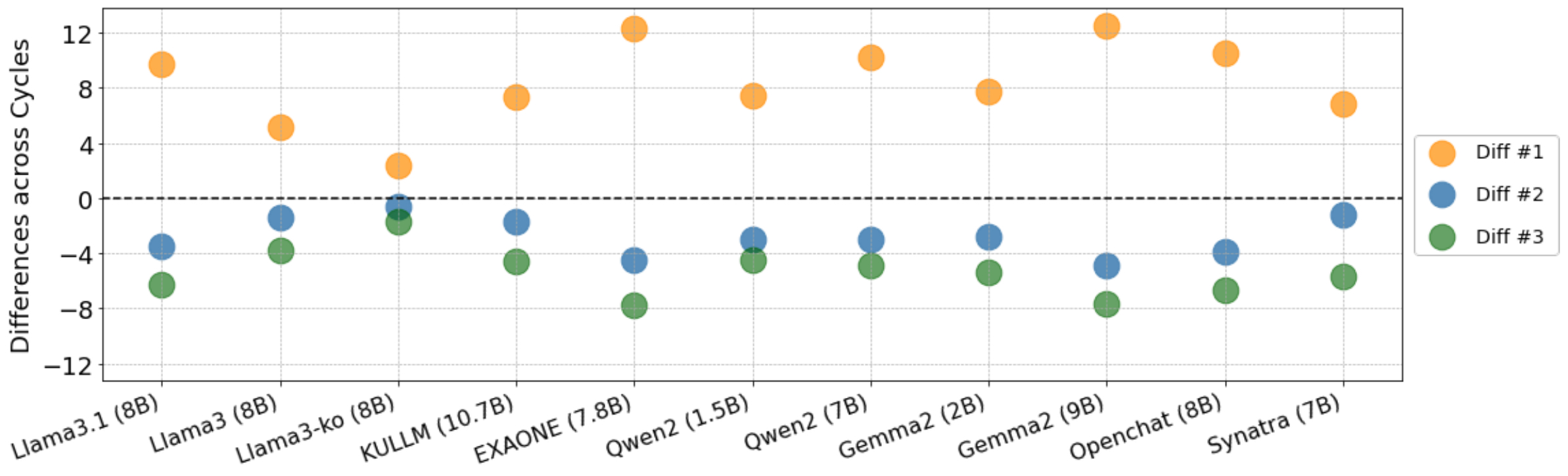}}
    \caption{Difference between the accuracy of each cycle and the average accuracy across all cycles after applying three rounds of cyclic permutation to the models. The further a circle is from the dashed line, the greater the deviation from the average, indicating greater inconsistency in the model's predictions. `Diff' in the legend means the form of the option order.}
    \label{figure_compare_cycles} 
\end{figure*}

We experimented with a diverse set of LLMs to assess their understanding of sentence endings, containing \texttt{Llama}-families, \texttt{Qwen2}, and \texttt{Gemma2} with parameter variations. We also selected Korean instruction-tuned models, including \texttt{KULLM3} and \texttt{EXAONE3}\footnote{Due to resource constraints, we conducted main experiments using models with up to 10.7B parameters, while the results of pilot experiments with a larger 70B model are presented in Appendix~\ref{appendix_c3}.}. The details regarding the models and metric are provided in Appendix~\ref{appendix_b2}.

\section{Discussion}

\textbf{How well do the models understand sentence endings?} The results of the sentence ending comprehension evaluation using the proposed dataset with the SE-\textit{always} task are presented in Table~\ref{table_SE_always}. We observed that their performance was relatively low, indicating that LLMs still have a limited understanding of Korean sentence endings\footnote{Although performance improved somewhat with the SE-\textit{absent} task in Table~\ref{table_SE_absent}, we observed that the overall performance level remained relatively low.}. To gain deeper insights into this situation, we compared the performance across several factors.

\subsection{Experimental Results}

\textbf{Which type of sentence ending form is more challenging?} We found that the accuracy for the \texttt{Imperative} forms was lower than that for the \texttt{Declarative} forms, indicating the greater difficulty in understanding sentence endings. This discrepancy likely arose because \texttt{Imperative} endings have more overlapping usage options than \texttt{Declarative} endings, making it more challenging for models to select contextually appropriate sentence endings.

\textbf{Does the contextual difficulty affect understanding of sentence endings?} We assumed that as the difficulty of the corpus increases, the models would struggle more to select the appropriate sentence endings. However, the results showed that corpus difficulty had a minimal effect on the accuracy of most models, except for \texttt{Gemma2} when predicting the usages of \texttt{Declarative} endings. This contrasts with the results in Table~\ref{table_human_agreement_scores}, which indicate that human annotation consistency decreased as corpus difficulty increased. It suggests that models faced more challenges in selecting the most natural sentence ending from the given options, regardless of the sentence's contextual complexity\footnote{Unlike in human annotation, the models were evaluated assuming no prior knowledge of specific usages, so we presented a broader range of options. While this may have influenced the results, the impact of difficulty on model accuracy during evaluation remained minimal.}.

\begin{table}[t!]
\centering
\small
\begin{tabular}{l|c|c|c}
\hline
Model (Parameters) & Diff \#1 & Diff \#2 & Diff \#3 \\ \hline
\texttt{Llama3.1} (8B)  & +9.69   & -3.47   & -6.22   \\
\texttt{Llama3} (8B)    & \underline{\smash{+5.15}}   & -1.39   & \underline{\smash{-3.75}}   \\
\texttt{Llama3-ko} (8B) & \textbf{+2.35}   & \textbf{-0.60}   & \textbf{-1.74}   \\
\texttt{KULLM3} (10.7B) & +7.39   & -1.67   & -4.54   \\
\texttt{EXAONE3} (7.8B) & +12.27  & -4.48   & -7.79   \\
\texttt{Qwen2} (1.5B)   & +7.46   & -2.99   & -4.47   \\
\texttt{Qwen2} (7B)     & +10.20  & -2.95   & -4.91   \\
\texttt{Gemma2} (2B)    & +7.78   & -2.76   & -5.40   \\
\texttt{Gemma2} (9B)    & +12.56  & -4.88   & -7.67   \\
\texttt{Openchat} (8B)  & +10.53  & -3.88   & -6.64   \\
\texttt{Synatra} (7B)   & +6.90   & \underline{\smash{-1.24}}   & -5.66   \\ \hline
\end{tabular}
\caption{Numeral differences between the accuracy of each cycle and the average accuracy of cyclic permutations. The top-2 smallest absolute differences in each cycle are highlighted in bold or underlined.}
\label{table_cycle_diff}
\end{table}

\begin{table*}[t!]
\begin{adjustbox}{max width=\textwidth}
\centering
\begin{tabular}{l|ccc|c|c|cc|cc|c|c}
\hline
\multirow{2}{*}{}                                                      \multirow{2}{*}{Sentence Endings}               & \texttt{Llama3.1} & \texttt{Llama3} & \texttt{Llama3-ko} & \texttt{KULLM3}           & \texttt{EXAONE3} & \multicolumn{2}{c|}{\texttt{Qwen2}} & \multicolumn{2}{c|}{\texttt{Gemma2}} & \texttt{Openchat}      & \texttt{Synatra} \\ \cline{2-12} 
                                                                                     & \multicolumn{3}{c|}{8B}                                  & 10.7B                     & 7.8B             & 1.5B             & 7B               & 2B                & 9B               & 8B                        & 7B                     \\ \hline
\multirow{3}{*}{\begin{tabular}[c]{@{}l@{}}\texttt{Declarative}\\ Forms\end{tabular}} & 16.58             & 17.70           & 22.58              & 20.89                     & 20.08            & 18.50            & 16.98            & 19.62             & 16.85            & 16.94                     & 18.39                  \\
                                                                                      & 14.39             & 18.63           & 23.27              & 21.02                     & 16.37            & 19.32            & 15.46            & 19.16             & 16.30            & 14.81                     & 18.45                  \\
                                                                                      & 14.70             & 17.90           & 21.94              & 21.32                     & 16.70            & 18.35            & 15.46            & 18.91             & 14.94            & 14.62                     & 17.36                  \\ \hline
Average                                                                               & 15.22             & 18.07           & \textbf{22.59}     & \underline{\smash{21.07}} & 17.71            & 18.72            & 15.96            & 19.23             & 16.03            & 15.45                     & 18.06                  \\ \hline
\multirow{3}{*}{\begin{tabular}[c]{@{}l@{}}\texttt{Imperative}\\ Forms\end{tabular}}  & 14.47             & 14.51           & 20.63              & 18.45                     & 20.96            & 14.63            & 16.52            & 17.30             & 13.96            & 20.29                     & 15.84                  \\
                                                                                      & 15.37             & 16.17           & 19.25              & 20.98                     & 19.43            & 16.06            & 17.84            & 16.91             & 16.44            & 20.09                     & 17.31                  \\
                                                                                      & 17.71             & 16.81           & 16.79              & 23.65                     & 21.86            & 17.22            & 20.08            & 19.60             & 19.00            & 21.20                     & 19.39                  \\ \hline
Average                                                                               & 15.85             & 15.82           & 18.88              & \textbf{21.02}            & 20.75            & 15.96            & 18.14            & 17.93             & 16.46            & \underline{\smash{20.52}} & 17.51                  \\ \hline
\end{tabular}
\end{adjustbox}
\caption{Accuracy of understanding Korean sentence endings across LLMs \underline{\smash{for the SE-\textit{absent} task}}. The method for determining final accuracy and the order of reported values by difficulty level match those presented in Table~\ref{table_SE_always}. The model with the highest average score across all models is highlighted in bold, whereas the second-best model is underlined.}
\label{table_SE_absent}
\end{table*}

\textbf{How does model parameter size affect understanding of sentence endings?} We compared the average accuracy based on the parameter count, in Figure~\ref{figure_compare_llms}. Although larger parameter counts in LLMs enhance performance in general tasks~\cite{wu2024performance, chowdhery2023palm}, our results showed that the parameter size had minimal impact. For instance, of the 11 models, \texttt{KULLM3} with the largest parameters ranked in the top 4 for both \texttt{Declarative} and \texttt{Imperative} ending predictions. Its performance was not significantly better than that of \texttt{Qwen2}, which had only 1.5B parameters. Similarly, \texttt{Gemma2}, with only 2B parameters, ranked in the top 2 in predicting \texttt{Declarative} endings. These relations suggest that all the models, regardless of the parameter count, face challenges in understanding Korean sentence endings.

\subsection{How does the option order of sentence endings affect the model's understanding?}

In our evaluation of sentence ending comprehension, we applied cyclic permutation~\cite{izacard2023atlas} to assess the impact of the order options on model predictions. While some models consistently predicted sentence endings accurately, regardless of the option order, most struggled to maintain performance despite minor changes due to cyclic permutation. The performance shift for each model is illustrated in Figure~\ref{figure_compare_cycles}.

The results showed that almost all models exhibited inconsistencies with cyclic permutation, regardless of the model type or parameter count. Notably, \texttt{EXAONE3} showed significant deviations, indicating poor robustness to changes in option order despite being additionally trained on a Korean dataset. Even larger models such as \texttt{KULLM3} and \texttt{Gemma2} (9B) were vulnerable to these shifts, indicating that even increased parameter sizes do not guarantee stability against changes in option order.

Conversely, \texttt{Llama3-ko} showed the smallest accuracy differences across cycles compared with that of the other models. It exhibited relatively greater consistency when compared with other models in the \texttt{Llama}-families and those with the same 8B parameters. Table~\ref{table_cycle_diff} provides a clear view of these differences, demonstrating that \texttt{Llama3-ko} had a significantly lower variability across cycles. It is likely due to the base model choice or the particular instruction-tuning approach, as opposed to other models trained on Korean datasets.

\begin{table}[t!]
\centering
\small
\begin{tabular}{l|c|c|c}
\hline
\begin{tabular}[c]{@{}l@{}}Model\\ (Parameters)\end{tabular} & \begin{tabular}[c]{@{}c@{}}SE-\textit{always}\\Task\end{tabular} & \begin{tabular}[c]{@{}c@{}}SE-\textit{absent}\\Task\end{tabular} & \begin{tabular}[c]{@{}c@{}}Increased\\Accuracy\end{tabular} \\ \hline
\texttt{Llama3.1} (8B)  & 10.77   & 15.53   & +4.76   \\
\texttt{Llama3} (8B)    & 13.63   & 16.94   & +3.30  \\
\texttt{Llama3-ko} (8B) & \textbf{15.00}   & \underline{\smash{20.73}}   & +5.73   \\
\texttt{KULLM3} (10.7B) & 13.71   & \textbf{21.04}   & \textbf{+7.33}   \\
\texttt{EXAONE3} (7.8B) & 12.31   & 19.23   & \underline{\smash{+6.92}}   \\
\texttt{Qwen2} (1.5B)   & 12.64   & 17.34   & +4.69   \\
\texttt{Qwen2} (7B)     & 11.47   & 17.05   & +5.57   \\
\texttt{Gemma2} (2B)    & 13.21   & 18.58   & +5.35   \\
\texttt{Gemma2} (9B)    & 12.16   & 16.24   & +4.08   \\
\texttt{Openchat} (8B)  & 11.89   & 17.98   & +6.09   \\
\texttt{Synatra} (7B)   & \underline{\smash{13.95}}   & 17.78   & +3.82   \\ \hline
\end{tabular}
\caption{Accuracy for both SE-\textit{always} and SE-\textit{absent} tasks, along with the improvements seen in the latter. These scores are averaged across all difficulty levels for both \texttt{Declarative} and \texttt{Imperative} forms. The top-2 highest scores in each column are highlighted in bold or underlined.}
\label{table_SE_all_acc}
\end{table}

\subsection{How does the possibility of no sentence ending affect the model's comprehension?}

The results from the SE-\textit{absent} task, in which the models were also given the \textit{no usages} option when evaluating sentence ending comprehension, are presented in Table~\ref{table_SE_absent}. All models exhibited a consistent performance improvement compared with that listed in Table~\ref{table_SE_always}, despite the increased number of samples used in the metric owing to the inclusion of the \textit{no usages} option. This suggests that all the models in our experiments, regardless of their model type, better understood sentence ending usage when accounting for the possibility that no valid usage exists.

Similar to the SE-\textit{always} task, we found that contextual difficulty had no significant impact on accuracy when predicting the usage of sentence endings in this task. This suggests that, regardless of the model's awareness of an absent sentence ending, the selection of the most natural usage is influenced more by the available options than by the context of the sentence.

In addition, when comparing model performance by parameter size, the largest model \texttt{KULLM3} ranked among the top 2 for both \texttt{Declarative} and \texttt{Imperative} forms. However, \texttt{Gemma2} (2B) outperformed the 9B models in all cases, suggesting that even with the awareness of missing sentence endings, the parameter size did not consistently improve the understanding of sentence endings.

We presented the average scores for both SE-\textit{always} and SE-\textit{absent} tasks, highlighting the improvements in the SE-\textit{absent} task in Table~\ref{table_SE_all_acc}. In general, the models performed better when informed of the possibility that no appropriate sentence ending might exist. Notably, models such as \texttt{KULLM3}, \texttt{Llama3-ko}, and \texttt{EXAONE3}, instruction-tuned with the Korean dataset exhibited a more significant performance boost, indicating that instruction tuning in Korean helps LLMs better grasp the nuances of sentence ending usage.

\section{Conclusion}

We proposed the Korean Sentence Endings (\textbf{KoSEnd}) dataset to evaluate the ability of various LLMs to understand the use of diverse Korean sentence endings, considering the language's agglutinative nature. The dataset was categorized into three difficulty levels to reflect the varying contextual nuances from different sources. We expanded all sentences with 15 types of sentence endings, including \texttt{Declarative} and \texttt{Imperative} forms, and applied a two-stage annotation process to label their natural usage.

By evaluating the performance of LLMs under two SE-\textit{always} and SE-\textit{absent} tasks, whether they were informed that a sentence ending might be absent, we found that models such as \texttt{Llama3-ko}, \texttt{Synatra}, and \texttt{KULLM3} achieved relatively high accuracy in both tasks. Furthermore, we examined performance variations based on the model parameters and the consistency of predictions through cyclic permutation. We observed that all models performed better when aware that a sentence ending might be missing. Moreover, the models instruction-tuned with a Korean dataset demonstrated strong prediction consistency and overall performance improvements.

Our study provides significant insights into evaluating linguistic knowledge in relatively low-resource agglutinative language, especially in Korean. Korean sentence endings convey not only grammatical roles but also semantic, emotional, and cultural nuances. Therefore, by using the proposed dataset, we expect to observe improvements in related performance for general tasks such as text generation, which we consider future work.

\section*{Limitations}

\textbf{The Risks of LLM-based Annotation} While we incorporated some human annotations to capture natural sentence ending usage, most samples were annotated using an LLM-based annotation, raising concerns about label quality and potential biases. To mitigate this, we conducted a pilot test as shown in Table~\ref{table_model_annotation_acc} to assess the reliability of this process. We further minimized bias by using human annotations as few-shot examples and employing cyclic permutation to reduce option order bias.

\textbf{Constraints on Task and Model Selection} We designed two tasks to evaluate each model's understanding of Korean sentence endings, but there remains ample room for further assessment using more diverse approaches. We aim to explore this comprehension from multiple angles, including its application to downstream tasks as future work.

Due to resource limitations, we focused on models with fewer parameters rather than larger 70B models, conducting an in-depth analysis to assess each model's understanding of Korean sentence endings from various perspectives. The pilot experiments with larger models can be found in Appendix~\ref{appendix_c3} in this aspect.

\section*{Ethics Statement}

Our proposed dataset comes from diverse sources with varying difficulty levels, which may lead to sentences that reflect biases or contain discriminatory language based on the nature of these corpora. As the proposed dataset focuses on expanding and annotating Korean sentence endings, we did not leverage potentially biased information from the original sources.

In our experiments to evaluate Korean sentence ending comprehension across various LLMs, there is a possibility that the inherent biases of the model could have influenced the predictions. We designed the task with multiple-choice questions to minimize such effects, focusing on the usage of each sentence ending. By framing this as a classification task and using greedy decoding, we aimed to avoid introducing additional biases from the models.

\section*{Acknowledgments}

This work was supported by the Institute of Information \& Communications Technology Planning \& Evaluation (IITP) grant funded by the Korea government (MSIT) [RS-2021-II211341, Artificial Intelligence Graduate School Program (Chung-Ang University)] and by the National Research Foundation of Korea (NRF) grant funded by the Korea government (MSIT) (RS-2025-00556246).

\bibliography{custom}

\begin{thebibliography}{59}
\providecommand{\natexlab}[1]{#1}

\bibitem[{Asai et~al.(2024)Asai, Kudugunta, Yu, Blevins, Gonen, Reid, Tsvetkov, Ruder, and Hajishirzi}]{asai-etal-2024-buffet}
Akari Asai, Sneha Kudugunta, Xinyan Yu, Terra Blevins, Hila Gonen, Machel Reid, Yulia Tsvetkov, Sebastian Ruder, and Hannaneh Hajishirzi. 2024.
\newblock \href {https://doi.org/10.18653/v1/2024.naacl-long.100} {{BUFFET}: Benchmarking large language models for few-shot cross-lingual transfer}.
\newblock In \emph{Proceedings of the 2024 Conference of the North American Chapter of the Association for Computational Linguistics: Human Language Technologies (Volume 1: Long Papers)}, pages 1771--1800, Mexico City, Mexico. Association for Computational Linguistics.

\bibitem[{Bai et~al.(2024)Bai, Lv, Zhang, Lyu, Tang, Huang, Du, Liu, Zeng, Hou, Dong, Tang, and Li}]{bai-etal-2024-longbench}
Yushi Bai, Xin Lv, Jiajie Zhang, Hongchang Lyu, Jiankai Tang, Zhidian Huang, Zhengxiao Du, Xiao Liu, Aohan Zeng, Lei Hou, Yuxiao Dong, Jie Tang, and Juanzi Li. 2024.
\newblock \href {https://aclanthology.org/2024.acl-long.172} {{L}ong{B}ench: A bilingual, multitask benchmark for long context understanding}.
\newblock In \emph{Proceedings of the 62nd Annual Meeting of the Association for Computational Linguistics (Volume 1: Long Papers)}, pages 3119--3137, Bangkok, Thailand. Association for Computational Linguistics.

\bibitem[{Brown et~al.(2020)Brown, Mann, Ryder, Subbiah, Kaplan, Dhariwal, Neelakantan, Shyam, Sastry, Askell et~al.}]{brown2020language}
Tom Brown, Benjamin Mann, Nick Ryder, Melanie Subbiah, Jared~D Kaplan, Prafulla Dhariwal, Arvind Neelakantan, Pranav Shyam, Girish Sastry, Amanda Askell, et~al. 2020.
\newblock Language models are few-shot learners.
\newblock \emph{Advances in neural information processing systems}, 33:1877--1901.

\bibitem[{Cahyawijaya et~al.(2023)Cahyawijaya, Lovenia, Aji, Winata, Wilie, Koto, Mahendra, Wibisono, Romadhony, Vincentio, Santoso, Moeljadi, Wirawan, Hudi, Wicaksono, Parmonangan, Alfina, Putra, Rahmadani, Oenang, Septiandri, Jaya, Dhole, Suryani, Putri, Su, Stevens, Nityasya, Adilazuarda, Hadiwijaya, Diandaru, Yu, Ghifari, Dai, Xu, Damapuspita, Wibowo, Tho, Karo~Karo, Fatyanosa, Ji, Neubig, Baldwin, Ruder, Fung, Sujaini, Sakti, and Purwarianti}]{cahyawijaya-etal-2023-nusacrowd}
Samuel Cahyawijaya, Holy Lovenia, Alham~Fikri Aji, Genta Winata, Bryan Wilie, Fajri Koto, Rahmad Mahendra, Christian Wibisono, Ade Romadhony, Karissa Vincentio, Jennifer Santoso, David Moeljadi, Cahya Wirawan, Frederikus Hudi, Muhammad~Satrio Wicaksono, Ivan Parmonangan, Ika Alfina, Ilham~Firdausi Putra, Samsul Rahmadani, Yulianti Oenang, Ali Septiandri, James Jaya, Kaustubh Dhole, Arie Suryani, Rifki~Afina Putri, Dan Su, Keith Stevens, Made~Nindyatama Nityasya, Muhammad Adilazuarda, Ryan Hadiwijaya, Ryandito Diandaru, Tiezheng Yu, Vito Ghifari, Wenliang Dai, Yan Xu, Dyah Damapuspita, Haryo Wibowo, Cuk Tho, Ichwanul Karo~Karo, Tirana Fatyanosa, Ziwei Ji, Graham Neubig, Timothy Baldwin, Sebastian Ruder, Pascale Fung, Herry Sujaini, Sakriani Sakti, and Ayu Purwarianti. 2023.
\newblock \href {https://doi.org/10.18653/v1/2023.findings-acl.868} {{N}usa{C}rowd: Open source initiative for {I}ndonesian {NLP} resources}.
\newblock In \emph{Findings of the Association for Computational Linguistics: ACL 2023}, pages 13745--13818, Toronto, Canada. Association for Computational Linguistics.

\bibitem[{Cahyawijaya et~al.(2024)Cahyawijaya, Lovenia, and Fung}]{cahyawijaya-etal-2024-llms}
Samuel Cahyawijaya, Holy Lovenia, and Pascale Fung. 2024.
\newblock \href {https://doi.org/10.18653/v1/2024.naacl-long.24} {{LLM}s are few-shot in-context low-resource language learners}.
\newblock In \emph{Proceedings of the 2024 Conference of the North American Chapter of the Association for Computational Linguistics: Human Language Technologies (Volume 1: Long Papers)}, pages 405--433, Mexico City, Mexico. Association for Computational Linguistics.

\bibitem[{Chowdhery et~al.(2023)Chowdhery, Narang, Devlin, Bosma, Mishra, Roberts, Barham, Chung, Sutton, Gehrmann et~al.}]{chowdhery2023palm}
Aakanksha Chowdhery, Sharan Narang, Jacob Devlin, Maarten Bosma, Gaurav Mishra, Adam Roberts, Paul Barham, Hyung~Won Chung, Charles Sutton, Sebastian Gehrmann, et~al. 2023.
\newblock Palm: Scaling language modeling with pathways.
\newblock \emph{Journal of Machine Learning Research}, 24(240):1--113.

\bibitem[{Ding et~al.(2023)Ding, Qin, Liu, Chia, Li, Joty, and Bing}]{ding-etal-2023-gpt}
Bosheng Ding, Chengwei Qin, Linlin Liu, Yew~Ken Chia, Boyang Li, Shafiq Joty, and Lidong Bing. 2023.
\newblock \href {https://doi.org/10.18653/v1/2023.acl-long.626} {Is {GPT}-3 a good data annotator?}
\newblock In \emph{Proceedings of the 61st Annual Meeting of the Association for Computational Linguistics (Volume 1: Long Papers)}, pages 11173--11195, Toronto, Canada. Association for Computational Linguistics.

\bibitem[{Ham et~al.(2020)Ham, Choe, Park, Choi, and Soh}]{ham-etal-2020-kornli}
Jiyeon Ham, Yo~Joong Choe, Kyubyong Park, Ilji Choi, and Hyungjoon Soh. 2020.
\newblock \href {https://doi.org/10.18653/v1/2020.findings-emnlp.39} {{K}or{NLI} and {K}or{STS}: New benchmark datasets for {K}orean natural language understanding}.
\newblock In \emph{Findings of the Association for Computational Linguistics: EMNLP 2020}, pages 422--430, Online. Association for Computational Linguistics.

\bibitem[{Hayes and Krippendorff(2007)}]{hayes2007answering}
Andrew~F Hayes and Klaus Krippendorff. 2007.
\newblock Answering the call for a standard reliability measure for coding data.
\newblock \emph{Communication methods and measures}, 1(1):77--89.

\bibitem[{He et~al.(2024)He, Lin, Gong, Jin, Zhang, Lin, Jiao, Yiu, Duan, and Chen}]{he-etal-2024-annollm}
Xingwei He, Zhenghao Lin, Yeyun Gong, A-Long Jin, Hang Zhang, Chen Lin, Jian Jiao, Siu~Ming Yiu, Nan Duan, and Weizhu Chen. 2024.
\newblock \href {https://doi.org/10.18653/v1/2024.naacl-industry.15} {{A}nno{LLM}: Making large language models to be better crowdsourced annotators}.
\newblock In \emph{Proceedings of the 2024 Conference of the North American Chapter of the Association for Computational Linguistics: Human Language Technologies (Volume 6: Industry Track)}, pages 165--190, Mexico City, Mexico. Association for Computational Linguistics.

\bibitem[{Hendrycks et~al.(2021)Hendrycks, Burns, Basart, Zou, Mazeika, Song, and Steinhardt}]{hendryckstest2021}
Dan Hendrycks, Collin Burns, Steven Basart, Andy Zou, Mantas Mazeika, Dawn Song, and Jacob Steinhardt. 2021.
\newblock Measuring massive multitask language understanding.
\newblock \emph{Proceedings of the International Conference on Learning Representations (ICLR)}.

\bibitem[{Huang et~al.(2023)Huang, Tang, Zhang, Zhao, Song, Xia, and Wei}]{huang-etal-2023-languages}
Haoyang Huang, Tianyi Tang, Dongdong Zhang, Xin Zhao, Ting Song, Yan Xia, and Furu Wei. 2023.
\newblock \href {https://doi.org/10.18653/v1/2023.findings-emnlp.826} {Not all languages are created equal in {LLM}s: Improving multilingual capability by cross-lingual-thought prompting}.
\newblock In \emph{Findings of the Association for Computational Linguistics: EMNLP 2023}, pages 12365--12394, Singapore. Association for Computational Linguistics.

\bibitem[{Hwang et~al.(2024)Hwang, Kim, Bae, Bang, Lee, and Jung}]{hwang2024kosmic}
Yerin Hwang, Yongil Kim, Hyunkyung Bae, Jeesoo Bang, Hwanhee Lee, and Kyomin Jung. 2024.
\newblock Kosmic: Korean text similarity metric reflecting honorific distinctions.
\newblock In \emph{Proceedings of the 2024 Joint International Conference on Computational Linguistics, Language Resources and Evaluation (LREC-COLING 2024)}, pages 9954--9960.

\bibitem[{Izacard et~al.(2023)Izacard, Lewis, Lomeli, Hosseini, Petroni, Schick, Dwivedi-Yu, Joulin, Riedel, and Grave}]{izacard2023atlas}
Gautier Izacard, Patrick Lewis, Maria Lomeli, Lucas Hosseini, Fabio Petroni, Timo Schick, Jane Dwivedi-Yu, Armand Joulin, Sebastian Riedel, and Edouard Grave. 2023.
\newblock Atlas: Few-shot learning with retrieval augmented language models.
\newblock \emph{Journal of Machine Learning Research}, 24(251):1--43.

\bibitem[{Jang et~al.(2022)Jang, Kim, Kwon, and Davis}]{jang-etal-2022-kobest}
Myeongjun Jang, Dohyung Kim, Deuk~Sin Kwon, and Eric Davis. 2022.
\newblock \href {https://aclanthology.org/2022.coling-1.325} {{K}o{BEST}: {K}orean balanced evaluation of significant tasks}.
\newblock In \emph{Proceedings of the 29th International Conference on Computational Linguistics}, pages 3697--3708, Gyeongju, Republic of Korea. International Committee on Computational Linguistics.

\bibitem[{Jang et~al.(2024)Jang, Lee, and Yu}]{jang-etal-2024-kodialogbench}
Seongbo Jang, Seonghyeon Lee, and Hwanjo Yu. 2024.
\newblock \href {https://aclanthology.org/2024.lrec-main.865} {{K}o{D}ialog{B}ench: Evaluating conversational understanding of language models with {K}orean dialogue benchmark}.
\newblock In \emph{Proceedings of the 2024 Joint International Conference on Computational Linguistics, Language Resources and Evaluation (LREC-COLING 2024)}, pages 9905--9925, Torino, Italia. ELRA and ICCL.

\bibitem[{Jin et~al.(2024)Jin, Kim, Lee, Yoo, Oh, and Lee}]{jin2024kobbq}
Jiho Jin, Jiseon Kim, Nayeon Lee, Haneul Yoo, Alice Oh, and Hwaran Lee. 2024.
\newblock Kobbq: Korean bias benchmark for question answering.
\newblock \emph{Transactions of the Association for Computational Linguistics}, 12:507--524.

\bibitem[{Kaya and Tantu{\u{g}}(2024)}]{kaya2024effect}
Yi{\u{g}}it~Bekir Kaya and A~C{\"u}neyd Tantu{\u{g}}. 2024.
\newblock Effect of tokenization granularity for turkish large language models.
\newblock \emph{Intelligent Systems with Applications}, 21:200335.

\bibitem[{Ke et~al.(2020)Ke, Ji, Liu, Zhu, and Huang}]{ke-etal-2020-sentilare}
Pei Ke, Haozhe Ji, Siyang Liu, Xiaoyan Zhu, and Minlie Huang. 2020.
\newblock \href {https://doi.org/10.18653/v1/2020.emnlp-main.567} {{S}enti{LARE}: Sentiment-aware language representation learning with linguistic knowledge}.
\newblock In \emph{Proceedings of the 2020 Conference on Empirical Methods in Natural Language Processing (EMNLP)}, pages 6975--6988, Online. Association for Computational Linguistics.

\bibitem[{Kim et~al.(2024{\natexlab{a}})Kim, Suk, Oh, Yoo, Thorne, and Oh}]{kim-etal-2024-click}
Eunsu Kim, Juyoung Suk, Philhoon Oh, Haneul Yoo, James Thorne, and Alice Oh. 2024{\natexlab{a}}.
\newblock \href {https://aclanthology.org/2024.lrec-main.296} {{CLI}c{K}: A benchmark dataset of cultural and linguistic intelligence in {K}orean}.
\newblock In \emph{Proceedings of the 2024 Joint International Conference on Computational Linguistics, Language Resources and Evaluation (LREC-COLING 2024)}, pages 3335--3346, Torino, Italia. ELRA and ICCL.

\bibitem[{Kim et~al.(2024{\natexlab{b}})Kim, Lee, Han, Choi, and Jung}]{kim2024does}
Jong~Myoung Kim, Young-Jun Lee, Yong-Jin Han, Ho-Jin Choi, and Sangkeun Jung. 2024{\natexlab{b}}.
\newblock \href {https://openreview.net/forum?id=yfyHxvVzZT} {Does incomplete syntax influence korean language model? focusing on word order and case markers}.
\newblock In \emph{First Conference on Language Modeling}.

\bibitem[{Kurihara et~al.(2022)Kurihara, Kawahara, and Shibata}]{kurihara2022jglue}
Kentaro Kurihara, Daisuke Kawahara, and Tomohide Shibata. 2022.
\newblock Jglue: Japanese general language understanding evaluation.
\newblock In \emph{Proceedings of the Thirteenth Language Resources and Evaluation Conference}, pages 2957--2966.

\bibitem[{Kwon et~al.(2023)Kwon, Li, Zhuang, Sheng, Zheng, Yu, Gonzalez, Zhang, and Stoica}]{kwon2023efficient}
Woosuk Kwon, Zhuohan Li, Siyuan Zhuang, Ying Sheng, Lianmin Zheng, Cody~Hao Yu, Joseph~E. Gonzalez, Hao Zhang, and Ion Stoica. 2023.
\newblock Efficient memory management for large language model serving with pagedattention.
\newblock In \emph{Proceedings of the ACM SIGOPS 29th Symposium on Operating Systems Principles}.

\bibitem[{Lab and research(2023)}]{kullm}
NLP \&~AI Lab and Human-Inspired~AI research. 2023.
\newblock Kullm: Korea university large language model project.
\newblock \url{https://github.com/nlpai-lab/kullm}.

\bibitem[{Lee(2005)}]{lee2005korean}
Iksop Lee. 2005.
\newblock \emph{Korean Grammar}, volume~33.
\newblock Seoul National University Press.

\bibitem[{Lee et~al.(2024)Lee, Kim, Kim, Kim, Won, Lee, and Choi}]{lee-etal-2024-kornat}
Jiyoung Lee, Minwoo Kim, Seungho Kim, Junghwan Kim, Seunghyun Won, Hwaran Lee, and Edward Choi. 2024.
\newblock \href {https://aclanthology.org/2024.findings-acl.666} {{K}or{NAT}: {LLM} alignment benchmark for {K}orean social values and common knowledge}.
\newblock In \emph{Findings of the Association for Computational Linguistics ACL 2024}, pages 11177--11213, Bangkok, Thailand and virtual meeting. Association for Computational Linguistics.

\bibitem[{Leong et~al.(2023)Leong, Ngui, Susanto, Rengarajan, Sarveswaran, and Tjhi}]{leong2023bhasa}
Wei~Qi Leong, Jian~Gang Ngui, Yosephine Susanto, Hamsawardhini Rengarajan, Kengatharaiyer Sarveswaran, and William~Chandra Tjhi. 2023.
\newblock Bhasa: A holistic southeast asian linguistic and cultural evaluation suite for large language models.
\newblock \emph{arXiv preprint arXiv:2309.06085}.

\bibitem[{Li et~al.(2024)Li, Shi, Liu, Yang, Liu, and Du}]{li2024quantifying}
Zihao Li, Yucheng Shi, Zirui Liu, Fan Yang, Ninghao Liu, and Mengnan Du. 2024.
\newblock Quantifying multilingual performance of large language models across languages.
\newblock \emph{arXiv preprint arXiv:2404.11553}.

\bibitem[{Lim et~al.(2019)Lim, Kim, and Lee}]{lim2019korquad1}
Seungyoung Lim, Myungji Kim, and Jooyoul Lee. 2019.
\newblock Korquad1. 0: Korean qa dataset for machine reading comprehension.
\newblock \emph{arXiv preprint arXiv:1909.07005}.

\bibitem[{Limisiewicz et~al.(2023)Limisiewicz, Balhar, and Mare{\v{c}}ek}]{limisiewicz-etal-2023-tokenization}
Tomasz Limisiewicz, Ji{\v{r}}{\'\i} Balhar, and David Mare{\v{c}}ek. 2023.
\newblock \href {https://doi.org/10.18653/v1/2023.findings-acl.350} {Tokenization impacts multilingual language modeling: Assessing vocabulary allocation and overlap across languages}.
\newblock In \emph{Findings of the Association for Computational Linguistics: ACL 2023}, pages 5661--5681, Toronto, Canada. Association for Computational Linguistics.

\bibitem[{LiteLLM(2025)}]{litellm}
LiteLLM. 2025.
\newblock \href {https://docs.litellm.ai/docs/} {Litellm documentation}.
\newblock Accessed on February 1, 2025.

\bibitem[{Liu et~al.(2024)Liu, Zhang, Zhao, Luu, and Bing}]{liu2024translation}
Chaoqun Liu, Wenxuan Zhang, Yiran Zhao, Anh~Tuan Luu, and Lidong Bing. 2024.
\newblock Is translation all you need? a study on solving multilingual tasks with large language models.
\newblock \emph{arXiv preprint arXiv:2403.10258}.

\bibitem[{Maxutov et~al.(2024)Maxutov, Myrzakhmet, and Braslavski}]{maxutov2024llms}
Akylbek Maxutov, Ayan Myrzakhmet, and Pavel Braslavski. 2024.
\newblock Do llms speak kazakh? a pilot evaluation of seven models.
\newblock In \emph{Proceedings of the First Workshop on Natural Language Processing for Turkic Languages (SIGTURK 2024)}, pages 81--91.

\bibitem[{Meta(2024{\natexlab{a}})}]{meta2024llama3.1}
Meta. 2024{\natexlab{a}}.
\newblock \href {https://ai.meta.com/blog/meta-llama-3-1/} {Introducing llama 3.1: Our most capable models to date}.
\newblock Accessed on February 1, 2025.

\bibitem[{Meta(2024{\natexlab{b}})}]{meta2024llama3}
Meta. 2024{\natexlab{b}}.
\newblock \href {https://ai.meta.com/blog/meta-llama-3/} {Introducing meta llama 3: The most capable openly available llm to date}.
\newblock Accessed on February 1, 2025.

\bibitem[{Miaschi et~al.(2020)Miaschi, Brunato, Dell{'}Orletta, and Venturi}]{miaschi-etal-2020-linguistic}
Alessio Miaschi, Dominique Brunato, Felice Dell{'}Orletta, and Giulia Venturi. 2020.
\newblock \href {https://doi.org/10.18653/v1/2020.coling-main.65} {Linguistic profiling of a neural language model}.
\newblock In \emph{Proceedings of the 28th International Conference on Computational Linguistics}, pages 745--756, Barcelona, Spain (Online). International Committee on Computational Linguistics.

\bibitem[{Nasution and Onan(2024)}]{nasution2024chatgpt}
Arbi~Haza Nasution and Aytug Onan. 2024.
\newblock Chatgpt label: Comparing the quality of human-generated and llm-generated annotations in low-resource language nlp tasks.
\newblock \emph{IEEE Access}.

\bibitem[{OpenRouter(2025)}]{openrouter}
OpenRouter. 2025.
\newblock \href {https://openrouter.ai/docs/quickstart} {Openrouter documentation}.
\newblock Accessed on February 1, 2025.

\bibitem[{Park et~al.(2024{\natexlab{a}})Park, Kim, Kim, Cho, Kim, Lee, Kim, and Lee}]{park-etal-2024-open}
Chanjun Park, Hyeonwoo Kim, Dahyun Kim, SeongHwan Cho, Sanghoon Kim, Sukyung Lee, Yungi Kim, and Hwalsuk Lee. 2024{\natexlab{a}}.
\newblock \href {https://aclanthology.org/2024.acl-long.177} {Open {K}o-{LLM} leaderboard: Evaluating large language models in {K}orean with {K}o-h5 benchmark}.
\newblock In \emph{Proceedings of the 62nd Annual Meeting of the Association for Computational Linguistics (Volume 1: Long Papers)}, pages 3220--3234, Bangkok, Thailand. Association for Computational Linguistics.

\bibitem[{Park et~al.(2024{\natexlab{b}})Park, Lee, Jeong, Park, and Lee}]{park2024pragmatic}
Dojun Park, Jiwoo Lee, Hyeyun Jeong, Seohyun Park, and Sungeun Lee. 2024{\natexlab{b}}.
\newblock Pragmatic competence evaluation of large language models for korean.
\newblock \emph{arXiv preprint arXiv:2403.12675}.

\bibitem[{Park et~al.(2021)Park, Moon, Kim, Cho, Han, Park, Song, Kim, Song, Oh, Lee, Oh, Lyu, Jeong, Lee, Seo, Lee, Kim, Lee, Jang, Do, Kim, Lim, Lee, Park, Shin, Kim, Park, Oh, Ha~(NAVER AI~Lab), and Cho}]{park-etal-2021-klue}
Sungjoon Park, Jihyung Moon, Sungdong Kim, Won~Ik Cho, Ji~Yoon Han, Jangwon Park, Chisung Song, Junseong Kim, Youngsook Song, Taehwan Oh, Joohong Lee, Juhyun Oh, Sungwon Lyu, Younghoon Jeong, Inkwon Lee, Sangwoo Seo, Dongjun Lee, Hyunwoo Kim, Myeonghwa Lee, Seongbo Jang, Seungwon Do, Sunkyoung Kim, Kyungtae Lim, Jongwon Lee, Kyumin Park, Jamin Shin, Seonghyun Kim, Lucy Park, Alice Oh, Jung-Woo Ha~(NAVER AI~Lab), and Kyunghyun Cho. 2021.
\newblock \href {https://datasets-benchmarks-proceedings.neurips.cc/paper_files/paper/2021/file/98dce83da57b0395e163467c9dae521b-Paper-round2.pdf} {Klue: Korean language understanding evaluation}.
\newblock In \emph{Proceedings of the Neural Information Processing Systems Track on Datasets and Benchmarks}, volume~1.

\bibitem[{Petrov et~al.(2024)Petrov, La~Malfa, Torr, and Bibi}]{petrov2024language}
Aleksandar Petrov, Emanuele La~Malfa, Philip Torr, and Adel Bibi. 2024.
\newblock Language model tokenizers introduce unfairness between languages.
\newblock \emph{Advances in Neural Information Processing Systems}, 36.

\bibitem[{Rajpurkar et~al.(2018)Rajpurkar, Jia, and Liang}]{rajpurkar-etal-2018-know}
Pranav Rajpurkar, Robin Jia, and Percy Liang. 2018.
\newblock \href {https://doi.org/10.18653/v1/P18-2124} {Know what you don{'}t know: Unanswerable questions for {SQ}u{AD}}.
\newblock In \emph{Proceedings of the 56th Annual Meeting of the Association for Computational Linguistics (Volume 2: Short Papers)}, pages 784--789, Melbourne, Australia. Association for Computational Linguistics.

\bibitem[{Rajpurkar et~al.(2016)Rajpurkar, Zhang, Lopyrev, and Liang}]{rajpurkar-etal-2016-squad}
Pranav Rajpurkar, Jian Zhang, Konstantin Lopyrev, and Percy Liang. 2016.
\newblock \href {https://doi.org/10.18653/v1/D16-1264} {{SQ}u{AD}: 100,000+ questions for machine comprehension of text}.
\newblock In \emph{Proceedings of the 2016 Conference on Empirical Methods in Natural Language Processing}, pages 2383--2392, Austin, Texas. Association for Computational Linguistics.

\bibitem[{Research et~al.(2024)Research, An, Bae, Choi, Choi, Choi, Hong, Hong, Hwang, Jeon et~al.}]{research2024exaone}
LG~Research, Soyoung An, Kyunghoon Bae, Eunbi Choi, Stanley~Jungkyu Choi, Yemuk Choi, Seokhee Hong, Yeonjung Hong, Junwon Hwang, Hyojin Jeon, et~al. 2024.
\newblock Exaone 3.0 7.8 b instruction tuned language model.
\newblock \emph{arXiv preprint arXiv:2408.03541}.

\bibitem[{Seo et~al.(2024)Seo, Lee, Park, Hong, Lee, and Lim}]{seo2024kocommongen}
Jaehyung Seo, Jaewook Lee, Chanjun Park, SeongTae Hong, Seungjun Lee, and Heui-Seok Lim. 2024.
\newblock Kocommongen v2: A benchmark for navigating korean commonsense reasoning challenges in large language models.
\newblock In \emph{Findings of the Association for Computational Linguistics ACL 2024}, pages 2390--2415.

\bibitem[{Sohn(2001)}]{sohn2001korean}
Ho-Min Sohn. 2001.
\newblock \emph{The korean language}.
\newblock Cambridge University Press.

\bibitem[{Son et~al.(2024{\natexlab{a}})Son, Lee, Kim, Kim, Muennighoff, Choi, Park, Yoo, and Biderman}]{son2024kmmlu}
Guijin Son, Hanwool Lee, Sungdong Kim, Seungone Kim, Niklas Muennighoff, Taekyoon Choi, Cheonbok Park, Kang~Min Yoo, and Stella Biderman. 2024{\natexlab{a}}.
\newblock Kmmlu: Measuring massive multitask language understanding in korean.
\newblock \emph{arXiv preprint arXiv:2402.11548}.

\bibitem[{Son et~al.(2024{\natexlab{b}})Son, Lee, Kim, Kim, Lee, Yeom, Jung, Kim, and Kim}]{son-etal-2024-hae}
Guijin Son, Hanwool Lee, Suwan Kim, Huiseo Kim, Jae~cheol Lee, Je~Won Yeom, Jihyu Jung, Jung~woo Kim, and Songseong Kim. 2024{\natexlab{b}}.
\newblock \href {https://aclanthology.org/2024.lrec-main.704} {{HAE}-{RAE} bench: Evaluation of {K}orean knowledge in language models}.
\newblock In \emph{Proceedings of the 2024 Joint International Conference on Computational Linguistics, Language Resources and Evaluation (LREC-COLING 2024)}, pages 7993--8007, Torino, Italia. ELRA and ICCL.

\bibitem[{Song et~al.(2024)Song, Huang, Zhou, and Ma}]{song2024multilingual}
Jiayang Song, Yuheng Huang, Zhehua Zhou, and Lei Ma. 2024.
\newblock Multilingual blending: Llm safety alignment evaluation with language mixture.
\newblock \emph{arXiv preprint arXiv:2407.07342}.

\bibitem[{Team et~al.(2024)Team, Riviere, Pathak, Sessa, Hardin, Bhupatiraju, Hussenot, Mesnard, Shahriari, Ram{\'e} et~al.}]{team2024gemma}
Gemma Team, Morgane Riviere, Shreya Pathak, Pier~Giuseppe Sessa, Cassidy Hardin, Surya Bhupatiraju, L{\'e}onard Hussenot, Thomas Mesnard, Bobak Shahriari, Alexandre Ram{\'e}, et~al. 2024.
\newblock Gemma 2: Improving open language models at a practical size.
\newblock \emph{arXiv preprint arXiv:2408.00118}.

\bibitem[{Wang et~al.(2018)Wang, Singh, Michael, Hill, Levy, and Bowman}]{wang-etal-2018-glue}
Alex Wang, Amanpreet Singh, Julian Michael, Felix Hill, Omer Levy, and Samuel Bowman. 2018.
\newblock \href {https://doi.org/10.18653/v1/W18-5446} {{GLUE}: A multi-task benchmark and analysis platform for natural language understanding}.
\newblock In \emph{Proceedings of the 2018 {EMNLP} Workshop {B}lackbox{NLP}: Analyzing and Interpreting Neural Networks for {NLP}}, pages 353--355, Brussels, Belgium. Association for Computational Linguistics.

\bibitem[{Weissweiler et~al.(2023)Weissweiler, Hofmann, Kantharuban, Cai, Dutt, Hengle, Kabra, Kulkarni, Vijayakumar, Yu, Schuetze, Oflazer, and Mortensen}]{weissweiler-etal-2023-counting}
Leonie Weissweiler, Valentin Hofmann, Anjali Kantharuban, Anna Cai, Ritam Dutt, Amey Hengle, Anubha Kabra, Atharva Kulkarni, Abhishek Vijayakumar, Haofei Yu, Hinrich Schuetze, Kemal Oflazer, and David Mortensen. 2023.
\newblock \href {https://doi.org/10.18653/v1/2023.emnlp-main.401} {Counting the bugs in {C}hat{GPT}{'}s wugs: A multilingual investigation into the morphological capabilities of a large language model}.
\newblock In \emph{Proceedings of the 2023 Conference on Empirical Methods in Natural Language Processing}, pages 6508--6524, Singapore. Association for Computational Linguistics.

\bibitem[{Wilie et~al.(2020)Wilie, Vincentio, Winata, Cahyawijaya, Li, Lim, Soleman, Mahendra, Fung, Bahar, and Purwarianti}]{wilie-etal-2020-indonlu}
Bryan Wilie, Karissa Vincentio, Genta~Indra Winata, Samuel Cahyawijaya, Xiaohong Li, Zhi~Yuan Lim, Sidik Soleman, Rahmad Mahendra, Pascale Fung, Syafri Bahar, and Ayu Purwarianti. 2020.
\newblock \href {https://aclanthology.org/2020.aacl-main.85} {{I}ndo{NLU}: Benchmark and resources for evaluating {I}ndonesian natural language understanding}.
\newblock In \emph{Proceedings of the 1st Conference of the Asia-Pacific Chapter of the Association for Computational Linguistics and the 10th International Joint Conference on Natural Language Processing}, pages 843--857, Suzhou, China. Association for Computational Linguistics.

\bibitem[{Wu and Tang(2024)}]{wu2024performance}
Chuhan Wu and Ruiming Tang. 2024.
\newblock Performance law of large language models.
\newblock \emph{arXiv preprint arXiv:2408.09895}.

\bibitem[{Xiang et~al.(2022)Xiang, Li, Lian, Huang, Watanabe, and Liu}]{xiang-etal-2022-visualizing}
Jiannan Xiang, Huayang Li, Defu Lian, Guoping Huang, Taro Watanabe, and Lemao Liu. 2022.
\newblock \href {https://doi.org/10.18653/v1/2022.findings-acl.35} {Visualizing the relationship between encoded linguistic information and task performance}.
\newblock In \emph{Findings of the Association for Computational Linguistics: ACL 2022}, pages 410--422, Dublin, Ireland. Association for Computational Linguistics.

\bibitem[{Yang et~al.(2024)Yang, Yang, Hui, Zheng, Yu, Zhou, Li, Li, Liu, Huang et~al.}]{yang2024qwen2}
An~Yang, Baosong Yang, Binyuan Hui, Bo~Zheng, Bowen Yu, Chang Zhou, Chengpeng Li, Chengyuan Li, Dayiheng Liu, Fei Huang, et~al. 2024.
\newblock Qwen2 technical report.
\newblock \emph{arXiv preprint arXiv:2407.10671}.

\bibitem[{Yoon et~al.(2023)Yoon, Park, Kim, Cho, Park, Kim, Seo, and Oh}]{yoon-etal-2023-towards}
Soyoung Yoon, Sungjoon Park, Gyuwan Kim, Junhee Cho, Kihyo Park, Gyu~Tae Kim, Minjoon Seo, and Alice Oh. 2023.
\newblock \href {https://doi.org/10.18653/v1/2023.acl-long.371} {Towards standardizing {K}orean grammatical error correction: Datasets and annotation}.
\newblock In \emph{Proceedings of the 61st Annual Meeting of the Association for Computational Linguistics (Volume 1: Long Papers)}, pages 6713--6742, Toronto, Canada. Association for Computational Linguistics.

\bibitem[{Zhang et~al.(2023)Zhang, Aljunied, Gao, Chia, and Bing}]{zhang2023m3exam}
Wenxuan Zhang, Mahani Aljunied, Chang Gao, Yew~Ken Chia, and Lidong Bing. 2023.
\newblock M3exam: A multilingual, multimodal, multilevel benchmark for examining large language models.
\newblock \emph{Advances in Neural Information Processing Systems}, 36:5484--5505.

\end{thebibliography}

\newpage
\appendix

\section{\texorpdfstring{Further Details in\\\hspace*{1.75em}KoSEnd: Dataset Construction}{Further Details in KoSEnd: Dataset Construction}}
\label{appendix_a}

\subsection{Corpus Collection}
\label{appendix_a1}

We used the language learner corpora~\cite{yoon-etal-2023-towards} for the \texttt{Easy} corpus. We expected sentences from these less-proficient writers to contain simple vocabulary and more straightforward contexts. For the \texttt{Intermediate} and \texttt{Hard} corpus, we used a newspaper corpus from the National Institute of the Korean Language\footnote{Version 2023, \url{https://kli.korean.go.kr/corpus/request/corpusRegist.do\#none}} and summaries from academic papers\footnote{\url{https://www.aihub.or.kr/aihubdata/data/view.do?currMenu=115&topMenu=100&aihubDataSe=data&dataSetSn=90}}. We expected these texts to contain more complex vocabulary and fewer easily accessible contexts. Their information is presented in Table~\ref{table_corpus_info}.

We selected sentences that ended with verbs and adjectives, as these were suitable for expanding sentence endings. Sentences considered too short to provide adequate context for understanding sentence endings were excluded.

\begin{table}[h!]
\begin{adjustbox}{max width=\columnwidth}
\centering
\small
\begin{tabular}{l|lll}
\hline
Difficulty   & \multicolumn{3}{l}{Collected Sentences} \\ \hline
\texttt{Easy}         & \multicolumn{3}{l}{1,000 sentences \textit{from corrected Korean Learner Corpus}} \\ \hline
\texttt{Intermediate} & \multicolumn{3}{l}{\begin{tabular}[c]{@{}l@{}}1,000 sentences \textit{for each of the 9 news topics}\\ \;\;(IT and Science, Economy, Culture,\\ \;\;\;Beauty and Health, Society, Lifestyle,\\ \;\;\;Sports, Entertainment, Politics) \end{tabular}} \\ \hline
\texttt{Hard}         & \multicolumn{3}{l}{\begin{tabular}[c]{@{}l@{}}1,000 sentences \textit{for each of the 8 academic fields}\\ \;\;(Humanities, Agricultural and Marine Sciences,\\ \;\;\;Social Sciences, Interdisciplinary Studies,\\ \;\;\;Arts and Sports, Engineering,\\ \;\;\;Natural Sciences, Medicine and Pharmacy)\end{tabular}} \\ \hline
\end{tabular}
\end{adjustbox}
\caption{Corpus information for each difficulty level. For \texttt{Intermediate} and \texttt{Hard}, we ensured that the texts were gathered from diverse topics and fields.}
\label{table_corpus_info}
\end{table}

\subsection{Sentence Ending Expansion}
\label{appendix_a2}

In \texttt{Declarative} sentences, sentence endings such as the case \underline{\smash{(1)}} \{다, 는다, ㄴ다\} in Table~\ref{table_sentence_endings} can be used to convey different meanings such as \{\textit{statements}, \textit{exclamations}, \textit{questions}\}. The correct choice of sentence endings can vary depending on the reader's interpretation. For instance, ``최선을 다하\underline{\smash{으마}}'' is incorrect due to the verb stem form, while ``최선을 다하\underline{\smash{마}}'' is correct from the case \underline{\smash{(5)}}. However, sentences such as ``목숨을 바치는\underline{\smash{구나}}'' and ``목숨을 바치\underline{\smash{구나}}'' from the case \underline{\smash{(2)}} are both acceptable and cannot be considered incorrect. In this situation, we conducted a two-stage annotation process to label the most natural cases after expanding all the sentences.

\subsection{Two-stage Annotation}
\label{appendix_a3}

\textbf{Human Annotation} Three native Korean-speaking university graduates volunteered to this process. A single sentence can be expanded to 33 versions, using 15 different \texttt{Declarative} and \texttt{Imperative} forms, as outlined in Table~\ref{table_sentence_endings}. We asked them to annotate whether each expanded version was appropriate for the context in binary form. Within each set of 15 ending forms, there may be multiple valid labels, or none at all. For example, in the case \underline{\smash{(1)}} \{다, 는다, ㄴ다\}, there are three possible endings. Depending on the sentence context, anywhere from 0 to 3 of these endings may be considered appropriate. This labeling process is repeated for all forms from \underline{\smash{(1)}} to \underline{\smash{(15)}}. We especially noted that, depending on the context, \textit{there might be no single best option or several acceptable options}. In this context, we used majority voting for the results of the human annotation to determine the gold labels for each usage.

\textbf{LLM-based Annotation} We used the human annotation results as few-shot samples to label the remaining sentences. For example, when labeling sentences of the case \underline{\smash{(1)}}, we provided human-labeled examples of that form as 2-shot samples. We present the actual prompts used for LLM-based annotation as follows:

\begin{itemize}
    \item LLM-based annotation prompt in Korean
    
    \fbox{
          \begin{minipage}{0.8\columnwidth}  
            \small
            \# system\\
            당신은 한국어에 유능한 사람입니다. 당신의 업무는 종결 어미의 쓰임이 자연스러운 문장과 그에 부합하는 쓰임을 고르는 것입니다.\\
            \\
            \# user\\
            종결어미의 쓰임이 자연스러운 문장과 그에 부합하는 쓰임을 고르기 위해, 아래 예시를 참고할 수 있습니다.\\
            문장 보기: \{\textit{sentence\_sample1}\}\\
            쓰임 보기: \{\textit{usage\_sample1}\}\\
            \\
            문장 보기: \{\textit{sentence\_sample2}\}\\
            쓰임 보기: \{\textit{usage\_sample2}\}\\
            \\
            주어진 한국어 문장들 중 종결 어미의 쓰임이 자연스러운 문장을 고르세요. 자연스러운 문장은 여러 개일 수도 있고, 하나도 없을 수도 있습니다.\\
            문장 보기: \{\textit{sentence\_option}\}\\
            \\
            주어진 문장 쓰임 중 앞서 고른 문장에 제일 부합하는 것을 고르세요. 부합하는 쓰임은 여러 개일 수도 있고, 하나도 없을 수도 있습니다.\\
            쓰임 보기: \{\textit{usage\_option}\}\\
            \\
            문장 정답 및 쓰임 정답을 별도의 설명 없이 알파벳으로 골라주세요.
          \end{minipage}
    }

    \newpage
    \item translated in English
    
    \fbox{
          \begin{minipage}{0.8\columnwidth}  
            \small
            \# system\\
            You are fluent in Korean. Your task is to identify sentences where the sentence endings are naturally used and select the corresponding appropriate usage.\\
            \\
            \# user\\
            To determine the most natural sentence endings and their appropriate usage, you can refer to the examples below.\\
            Sentence options: \{\textit{sentence\_sample1}\}\\
            Usage options: \{\textit{usage\_sample1}\}\\
            \\
            Sentence options: \{\textit{sentence\_sample2}\}\\
            Usage options: \{\textit{usage\_sample2}\}\\
            \\
            From the given Korean sentences, select those with natural sentence endings. There may be multiple correct answers, or none at all.\\
            Sentence options: \{\textit{sentence\_option}\}\\
            \\
            Next, choose the usage option that best matches the selected sentence(s). Again, there may be multiple correct answers, or none at all.\\
            Usage options: \{\textit{usage\_option}\}\\
            \\
            Please provide your answers using the alphabet letter, without any additional explanation.
          \end{minipage}
    }
\end{itemize}

\section{\texorpdfstring{Further Details in\\\hspace*{1.75em}Sentence Ending Evaluation}{Further Details in Sentence Ending Evaluation}}
\label{appendix_b}

\subsection{Task Definition}
\label{appendix_b1}

In the two-stage annotation process, only specific candidates relevant to each usage were presented to the human annotators and models. For instance, options such as the case \underline{\smash{(1)}} \{다, 는다, ㄴ다\} and \underline{\smash{(2)}} \{구나, 는구나\} in Table~\ref{table_sentence_endings} were presented separately and not mixed. This approach ensured that, annotators or models could select the most appropriate sentence ending within that form, leading to the most natural choice for constructing the dataset.

In contrast, when evaluating the LLMs' understanding of sentence endings, we assumed that the model had no prior knowledge of the specific usage of the sentence. Thus, we combined options from all the forms and required the model to select the most natural sentence endings. To prevent the model from being influenced by the order of options, we applied cyclic permutation~\cite{izacard2023atlas}, expecting results would remain consistent regardless of the arrangement of options.

In the dataset construction process, sentences labeled as \textit{no usages}, indicating the absence of an ending across 15 possible cases of \texttt{Declarative} and \texttt{Imperative} endings, are detailed in Table~\ref{table_no_usages_info}.

\begin{table}[t!]
\begin{adjustbox}{max width=\columnwidth}
\centering
\small
\begin{tabular}{l|l|c|c}
\hline
Sentence Endings      & Difficulty            & \begin{tabular}[c]{@{}c@{}}\textit{no usages}\\Counts\end{tabular} & \begin{tabular}[c]{@{}c@{}}\textit{no usages}\\Ratio\end{tabular} \\ \hline
\multirow{3}{*}{\begin{tabular}[c]{@{}l@{}}\texttt{Declarative}\\ Forms\end{tabular}} & \texttt{Easy}         & 1,703       & 18.92\%     \\
                                                                                      & \texttt{Intermediate} & 568         & 6.31\%      \\
                                                                                      & \texttt{Hard}         & 1,379       & 15.32\%     \\ \hline
\multirow{3}{*}{\begin{tabular}[c]{@{}l@{}}\texttt{Imperative}\\ Forms\end{tabular}}  & \texttt{Easy}         & 3,149       & 52.48\%     \\
                                                                                      & \texttt{Intermediate} & 2,770       & 46.16\%     \\
                                                                                      & \texttt{Hard}         & 2,973       & 49.55\%     \\ \hline
\end{tabular}
\end{adjustbox}
\caption{Counts and proportions of sentences labeled as \textit{no usages} in the proposed dataset, categorized by sentence ending types and difficulty levels.}
\label{table_no_usages_info}
\end{table}

\begin{figure*}[t!]
    \centerline{\includegraphics[width=\textwidth]{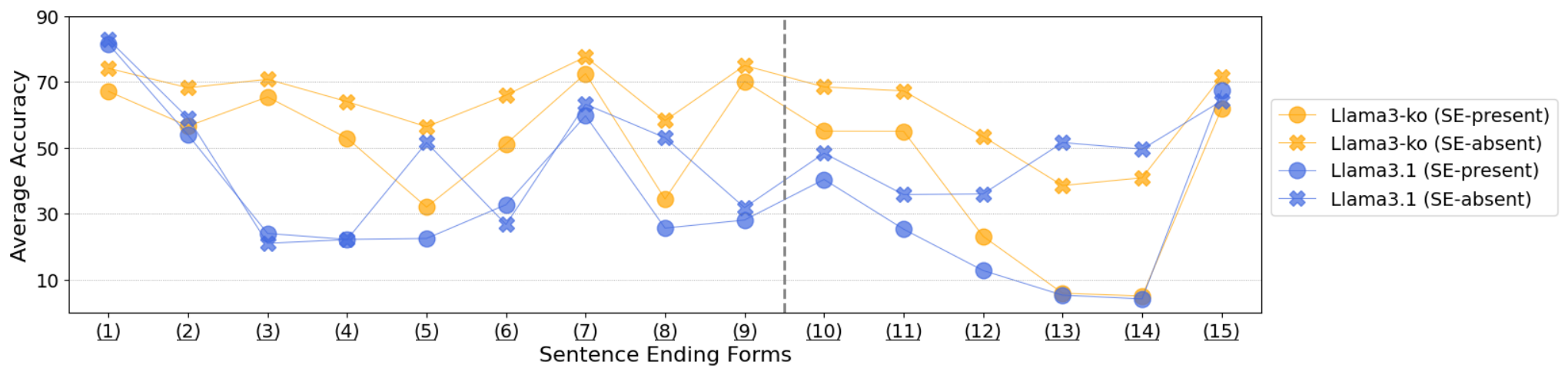}}
    \caption{Average scores for each sentence ending form of the two models, \texttt{Llama3-ko} and \texttt{Llama3.1}, which exhibited the best and worst performance in our experiments. The x-axis displays \underline{\smash{(1)}}–\underline{\smash{(9)}} for \texttt{Declarative} forms and \underline{\smash{(10)}}–\underline{\smash{(15)}} for \texttt{Imperative} forms, as shown in Table~\ref{table_sentence_endings}. These scores represent the average across all difficulty levels and cycles for each sentence ending form.}
    \label{figure_compare_each_endings} 
\end{figure*}

\subsection{Experimental Settings}
\label{appendix_b2}

The models to evaluate the understanding of Korean sentence endings are as follows: \texttt{Llama}-families~\cite{meta2024llama3.1, meta2024llama3}, \texttt{Gemma2}~\cite{team2024gemma}, and \texttt{Qwen2}~\cite{yang2024qwen2} were selected as the multilingual models. In addition, \texttt{KULLM3}~\cite{kullm} and \texttt{EXAONE3}~\cite{research2024exaone} were instruction-tuned using a Korean dataset. Specifically, as of September 2024, \texttt{Openchat}\footnote{\url{https://huggingface.co/openchat/openchat-3.6-8b-20240522}} and \texttt{Synatra}\footnote{\url{https://huggingface.co/maywell/Synatra-7B-v0.3-dpo}} were ranked as the top-2 models on the Open Ko-LLM Leaderboard\footnote{This leaderboard, a key benchmark for Korean language tasks using private test sets, features the top-performing models in Korean for various downstream tasks.}~\cite{park-etal-2024-open}. We set the temperature to 0 to enable greedy decoding for predicting the most natural usage of sentence endings. We used the vLLM library~\cite{kwon2023efficient} to enable efficient inference.

We measured accuracy by comparing the models' responses to the gold labels obtained through a two-stage annotation process. Each model generated responses to the same prompt three times using cyclic permutation, aligning with accuracy metrics from previous work~\cite{kim-etal-2024-click}.

\section{Further Details in Experiments}
\label{appendix_c}

\subsection{Post Processing}
\label{appendix_c1}

\begin{table}[t!]
\begin{adjustbox}{max width=\columnwidth}
\centering
\small
\begin{tabular}{l|l|c|c|c}
\hline
Tasks                                   & Model (Parameters)  & \texttt{Easy} & \texttt{Intermediate} & \texttt{Hard} \\ \hline
\multirow{4}{*}{\begin{tabular}[c]{@{}l@{}}SE-\textit{always}\\Task\end{tabular}} & \texttt{EXAONE3} (7.8B)& 0.013\%        & -                     & -             \\
                                   & \texttt{Qwen2} (7B)    & -             & -                     & 0.002\%        \\
                                   & \texttt{Gemma2} (9B)   & -             & -                     & 0.002\%        \\
                                   & \texttt{Synatra} (7B)  & 0.006\%        & -                     & -             \\ \hline
\multirow{3}{*}{\begin{tabular}[c]{@{}l@{}}SE-\textit{absent}\\Task\end{tabular}}   & \texttt{KULLM3} (10.7B)& 0.002\%        & 0.008\%                & 0.04\%         \\
                                   & \texttt{EXAONE3} (7.8B)& 0.02\%         & -                     & -             \\
                                   & \texttt{Synatra} (7B)  & -             & 0.004\%                & 0.002\%        \\ \hline
\end{tabular}
\end{adjustbox}
\caption{Hallucination rates for each task, based on the selected models. Any values not listed in the table were not classified as hallucinations according to our post-processing process.}
\label{table_hallucinations}
\end{table}

When we instructed the models, some generated additional explanations alongside their selections. To refine these outputs, we applied post-processing, prioritizing the alphabet following phrases like `correct answer’ or removing irrelevant characters that did not represent the answer. If the answer remained unclear after this process, we classified it as a hallucination. The hallucination rates for each model are shown in Table~\ref{table_hallucinations}. We excluded these hallucination samples from the evaluation.

\subsection{\texorpdfstring{Experimental Results\\\hspace*{2.5em}on Each Sentence Ending Form}{Experimental Results on Each Sentence Ending Form}}
\label{appendix_c2}

To analyze the impact of each sentence ending form on model performance, we reported the results for each form individually in Figure~\ref{figure_compare_each_endings}. Based on the results in Table~\ref{table_SE_all_acc}, we selected \texttt{Llama3-ko} and \texttt{Llama3.1}, which exhibited the highest and lowest performance in both SE-\textit{always} and SE-\textit{absent} tasks, respectively. In most cases, regardless of the sentence ending form, we observed performance improvements when the models were informed about the potential absence of a sentence ending. This trend was consistent across both \texttt{Llama3-ko} and \texttt{Llama3.1}, suggesting that recognizing the possibility of a missing sentence ending enhances their understanding of Korean sentence endings. 

Although \texttt{Llama3-ko} demonstrated strong performance across most sentence-ending forms, we observed that \texttt{Llama3.1} either outperformed or achieved comparable results to \texttt{Llama3-ko} in cases \underline{\smash{(1)}} and \underline{\smash{(13)}}\textasciitilde\underline{\smash{(15)}}. Cases \underline{\smash{(1)}}, \underline{\smash{(13)}}, and \underline{\smash{(14)}} represent the most commonly used forms, including \textit{statements} and \textit{informal speeches}. \texttt{Llama3.1}'s improved performance can be attributed to its training on larger dataset as a more recent model. Case \underline{\smash{(15)}} from the \texttt{Imperative} forms includes six different usages, the highest number of usages for any sentence ending form. This suggests that \texttt{Llama3.1}'s ability to handle a broader range of variations allowed it to perform comparably to \texttt{Llama3-ko}.

\subsection{Pilot Experiments with Larger Models}
\label{appendix_c3}

We conducted pilot experiments to evaluate Korean sentence endings using larger models not included in the main analysis. Due to time and budget constraints, we tested 1,000 samples for each combination of \texttt{Declarative} and \texttt{Imperative} sentence ending forms and three difficulty levels. We evaluated the \texttt{Llama3.1} 70B and \texttt{Qwen2.5} 72B models on SE-\textit{absent} task, using LiteLLM~\cite{litellm} and OpenRouter~\cite{openrouter}.

\begin{table}[t!]
\centering
\begin{adjustbox}{max width=\columnwidth}
\small
\begin{tabular}{l|cc|cc}
\hline
\multirow{2}{*}{Sentence Endings}                                                            & \multicolumn{2}{c|}{\texttt{Llama3.1}} & \multicolumn{2}{c}{\texttt{Qwen2.5}} \\ \cline{2-5} 
                                                                             & 8B            & 70B           & 7B           & 72B          \\ \hline
\multirow{3}{*}{\begin{tabular}[c]{@{}l@{}}\texttt{Declarative}\\ Forms\end{tabular}} & 29.90             & 39.30 (+31.43\%)             & 34.90            & 42.10 (+20.63\%)            \\
                                                                             & 26.60             & 34.60 (+30.07\%)             & 27.10            & 32.40 (+19.55\%)            \\
                                                                             & 26.10             & 31.60 (+21.07\%)             & 27.40            & 32.00 (+16.78\%)            \\ \hline
Average                                                                      & 27.53             & \underline{\smash{35.16}} (+27.71\%)             & 29.80            & \textbf{35.50} (+19.12\%)            \\ \hline
\multirow{3}{*}{\begin{tabular}[c]{@{}l@{}}\texttt{Imperative}\\ Forms\end{tabular}}  & 25.80             & 35.30 (+36.82\%)             & 45.80            & 49.60 (+8.29\%)            \\
                                                                             & 27.50             & 37.00 (+34.54\%)             & 45.00            & 48.80 (+8.44\%)            \\
                                                                             & 30.10             & 40.60 (+34.88\%)            & 49.30           & 53.40 (+8.31\%)           \\ \hline
Average                                                                      & 27.80            & 37.63 (+35.35\%)            & \underline{\smash{46.69}}           & \textbf{50.60} (+8.37\%)           \\ \hline
\end{tabular}
\end{adjustbox}
\caption{Accuracy of understanding Korean sentence endings with larger models \underline{\smash{for the SE-\textit{absent} task}} using 1,000 samples. For both \texttt{Declarative} and \texttt{Imperative} forms, the three reported values from the top represent results for \texttt{Easy}, \texttt{Intermediate}, and \texttt{Hard}, respectively. The top-2 highest averaged scores in each form are highlighted in bold or underlined. The values in parentheses represent the rate of performance improvement.}
\label{table_SE_absent_larger}
\end{table}

The results in Table~\ref{table_SE_absent_larger} showed that the larger models consistently outperformed smaller ones across all cases, regardless of sentence ending form or difficulty level. While larger models demonstrated capabilities in understanding sentence endings, the performance gains did not scale proportionally with the increase in parameter size. This indicates that even the larger models still face challenges in grasping the nuances of the Korean sentence endings.

Notably, \texttt{Qwen2.5} 7B demonstrated a relatively higher understanding of \texttt{Imperative} forms, even surpassing the performance of \texttt{Llama3.1} 70B. In contrast, within the \texttt{Llama3.1}-families, larger models consistently outperformed smaller ones with performance gains of around 30\%. This suggests that while \texttt{Llama3.1} showed greater improvements with larger model size, \texttt{Qwen2.5} achieved higher overall performance. This pilot experiments provided a broader perspective on the impact of our dataset. We hope these observations will help inform strategic decisions on model selection—both in terms of type and size—when assessing the understanding of Korean sentence endings.

\end{document}